\documentclass[lettersize,journal]{IEEEtran}
\usepackage{svg}

\usepackage{textcomp}
\usepackage{stfloats}

\usepackage{cite}

\usepackage{graphicx}
\usepackage{amsmath}
\usepackage{amssymb}
\usepackage{algorithm}
\usepackage{mathtools}

\usepackage[caption=false,font=normalsize,labelfont=sf,textfont=sf]{subfig}

\usepackage{times}
\usepackage{epsfig}

\usepackage{color}
\usepackage{xcolor}
\usepackage{booktabs,siunitx} 
\usepackage{makecell}
\usepackage{enumitem}
\usepackage[percent]{overpic}
\usepackage[caption=false]{subfig}
\usepackage{rotating}
\usepackage[misc]{ifsym}
\usepackage{xspace}
\usepackage{pifont}
\usepackage{multirow}
\usepackage{bm}
\usepackage[accsupp]{axessibility} 
\usepackage{tabularx}
\usepackage[noend]{algpseudocode}

\usepackage[pagebackref=true,breaklinks=true,colorlinks,bookmarks=false]{hyperref}
\usepackage[capitalize]{cleveref}

\usepackage{microtype}
\Crefname{section}{Section}{Sections}
\Crefname{table}{Table}{Tables}
\crefname{section}{Sec.}{Secs.}
\crefname{table}{Tab.}{Tabs.}

\if@mathematic
   
\else
   
\fi

\newcommand{\blue}[1]{\textcolor{black}{#1}}
\newcommand{\bluereb}[1]{\textcolor{black}{#1}}
\colorlet{bluereb}{black}

\newcommand{\ie}{\textit{i}.\textit{e}.\ }
\newcommand{\eg}{\textit{e}.\textit{g}.\ }
\newcommand{\wrt}{\textit{w}.\textit{r}.\textit{t}.\ }
\newcommand{\vs}{\textit{vs}.\ }

\newcommand{\best}[1]{\mathbf{#1}}
\newcommand{\scnd}[1]{\underline{#1}}

\newcommand{\PAR}[1]{\vskip4pt \noindent{\bf #1~}}

\newcommand*{\ourmodel}{UniDepthV2\@\xspace}
\newcommand{\cmark}{\ding{51}}
\newcommand{\xmark}{\ding{55}}

\begin{document}

\title{UniDepthV2:\\Universal Monocular Metric Depth Estimation\\Made Simpler}

\author{
Luigi~Piccinelli,
Christos~Sakaridis,
Yung-Hsu~Yang,
Mattia~Segu,
Siyuan~Li,
Wim~Abbeloos,
and~Luc~Van~Gool
\IEEEcompsocitemizethanks{
\IEEEcompsocthanksitem L.~Piccinelli, C.~Sakaridis, Y-H..~Yang, M.~Segu, and S.~Li are with ETH Z\"urich, Switzerland.
\IEEEcompsocthanksitem W.~Abbeloos is with Toyota Motor Europe, Belgium.
\IEEEcompsocthanksitem L.~Van Gool is with ETH Z\"urich, Switzerland, and with INSAIT, Sofia University, Bulgaria.}
}

%
%



\maketitle

\begin{abstract}
    Accurate monocular metric depth estimation (MMDE) is crucial to solving downstream tasks in 3D perception and modeling.
    However, the remarkable accuracy of recent MMDE methods is confined to their training domains.
    These methods fail to generalize to unseen domains even in the presence of moderate domain gaps, which hinders their practical applicability.
    We propose a new model, \ourmodel, capable of reconstructing metric 3D scenes from solely single images across domains.
    Departing from the existing MMDE paradigm, \ourmodel directly predicts metric 3D points from the input image at inference time without any additional information, striving for a universal and flexible MMDE solution.
    In particular, \ourmodel implements a self-promptable camera module predicting a dense camera representation to condition depth features.
    Our model exploits a pseudo-spherical output representation, which disentangles the camera and depth representations.
    In addition, we propose a geometric invariance loss that promotes the invariance of camera-prompted depth features.
    \ourmodel improves its predecessor UniDepth model via a new edge-guided loss which enhances the localization and sharpness of edges in the metric depth outputs, a revisited, simplified and more efficient architectural design, and an additional uncertainty-level output which enables downstream tasks requiring confidence.
    Thorough evaluations on ten depth datasets in a zero-shot regime consistently demonstrate the superior performance and generalization of \ourmodel.
    Code and models are available at: \href{https://github.com/lpiccinelli-eth/unidepth}{github.com/lpiccinelli-eth/UniDepth}.
\end{abstract}
    

\section{Introduction}
\label{sec:intro}

\IEEEPARstart{P}{recise} pixel-wise depth estimation is crucial to understanding the geometric scene structure, with applications in 3D modeling~\cite{deng2022nerf}, robotics~\cite{Zhou2019, dong2022depth4robotics}, and autonomous vehicles~\cite{wang2019depth4vehicles, park2021dd3d}.
However, delivering reliable metric scaled depth outputs is necessary to perform 3D reconstruction effectively, thus motivating the challenging and inherently ill-posed task of Monocular Metric Depth Estimation (MMDE).

\begin{figure}[ht]
    \centering
    \includegraphics[width=1.0\linewidth]{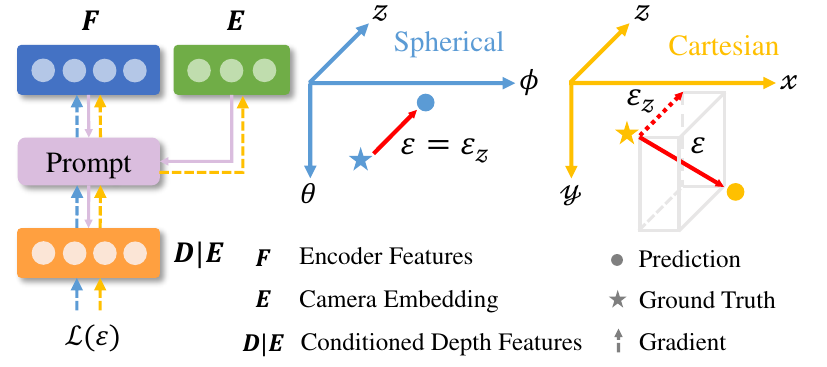}
    \vspace{-2em}
    \caption{
    We introduce \ourmodel, a novel approach that directly predicts 3D points in a scene with only one image as input. 
    \ourmodel incorporates a camera self-prompting mechanism and leverages a spherical 3D output space defined by azimuth and elevation angles, and depth($\theta$, $\phi$, $z$).
    This design effectively separates camera and depth optimization by avoiding gradient flowing to the camera module due to depth-related error ($\varepsilon_z$) compared to the standard Cartesian representation.}
    \label{fig:teaser}
\end{figure}

While existing MMDE methods~\cite{Eigen2014, Fu2018Dorn, Bhat2020adabins, Ranftl2021dpt, Patil2022p3depth, Yuan2022newcrf, piccinelli2023idisc} have demonstrated remarkable accuracy across different benchmarks, they require training and testing on datasets with similar camera intrinsics and scene scales.
Moreover, the training datasets typically have a limited size and contain little diversity in scenes and cameras.
These characteristics result in poor generalization to real-world inference scenarios~\cite{Wang2020traingermany}, where images are captured in uncontrolled, arbitrarily structured environments and cameras with arbitrary intrinsics. \blue{What makes the situation even worse is the imperfect nature of actual ground-truth depth which is used to supervise MMDE models, namely its sparsity and its incompleteness near edges, which results in blurry predictions with inaccurate fine-grained geometric details.}

Only a few methods~\cite{yin2023metric3d, guizilini2023zerodepth, hu2024metric3dv2} have addressed the challenging task of generalizable MMDE.
However, these methods assume controlled setups at test time, including camera intrinsics. 
While this assumption simplifies the task, it has two notable drawbacks.
Firstly, it does not address the full application spectrum, \eg in-the-wild video processing and crowd-sourced image analysis.
Secondly, the inherent camera parameter noise is directly injected into the model, leading to large inaccuracies in the high-noise case.

In this work, we address the more demanding task of generalizable MMDE \emph{without} any reliance on additional external information, such as camera parameters, thus defining the universal MMDE task.
Our approach, named \ourmodel{}, extends UniDepth~\cite{piccinelli2024unidepth} and is the first that attempts to solve this challenging task without restrictions on scene composition and setup and distinguishes itself through its general and adaptable nature. 
Unlike existing methods, \ourmodel delivers metric 3D predictions for any scene \emph{solely} from a single image, waiving the need for extra information about scene or camera.
Furthermore, \ourmodel flexibly allows for the incorporation of additional camera information at test time. \blue{Simultaneously, \ourmodel achieves sharper depth predictions with better-localized depth discontinuities than the original UniDepth model thanks to a novel edge-guided loss that enhances the consistency of the local structure of depth predictions around edges with the respective structure in the ground truth.}

\blue{The design of \ourmodel} introduces a camera module that outputs a non-parametric, \ie{}dense camera representation, serving as the prompt to the depth module. 
However, relying only on this single additional module clearly results in challenges related to training stability and scale ambiguity.
We propose an effective pseudo-spherical representation of the output space to disentangle the camera and depth dimensions of this space.
This representation employs azimuth and elevation angle components for the camera and a radial component for the depth, forming a perfect orthogonal space between the camera plane and the depth axis.
Moreover, \blue{the pinhole-based camera representation is positionally encoded via a sine encoding in \ourmodel, leading to a substantially more efficient computation compared to the spherical harmonic encoding of the pinhole-based representation of the original UniDepth.}
Figure~\ref{fig:teaser} depicts our camera self-prompting mechanism and the output space.
Additionally, we introduce a geometric invariance loss to enhance the robustness of depth estimation. 
The underlying idea is that the camera-conditioned depth \blue{outputs} from two views of the same image should exhibit reciprocal consistency.
In particular, we sample two geometric augmentations, creating different views for each training image, thus simulating different apparent cameras for the original scene. \blue{Besides the aforementioned consistency-oriented invariance loss, \ourmodel features an additional uncertainty output and respective loss. These pixel-level uncertainties are supervised with the differences between the respective depth predictions and their corresponding ground-truth values, and enable the utilization of our MMDE model in downstream tasks such as control which require confidence-aware perception inputs~\cite{bonzanini2021perception,mesbah2016stochastic,yang2023safe,bemporad2007robust} for certifiability.}

\blue{The overall contributions of the present, extended journal version of our work are the first universal MMDE methods, the original UniDepth and the newer \ourmodel,} which predict a point in metric 3D space for each pixel without \emph{any} input other than a single image. \blue{An earlier version of this work has appeared in the Conference on Computer Vision and Pattern Recognition~\cite{piccinelli2024unidepth} and has introduced our original UniDepth model. In~\cite{piccinelli2024unidepth}, we have first designed} a promptable camera module, an architectural component that learns a dense camera representation and allows for non-parametric camera conditioning.
Second, we \blue{have proposed} a pseudo-spherical representation of the output space, thus solving the intertwined nature of camera and depth prediction.
In addition, we \blue{have introduced} a geometric invariance loss to disentangle the camera information from the underlying 3D geometry of the scene.
\blue{Moreover, in the conference version, we have extensively evaluated and compared UniDepth}
on ten different datasets in a fair and comparable zero-shot setup to lay the ground for \blue{our novel} generalized MMDE task.
Owing to its design, UniDepth consistently set the state of the art even compared with non-zero-shot methods, ranking first \blue{at the time of its appearance} in the competitive official KITTI Depth Prediction Benchmark.
\blue{Compared to the aforementioned conference version, this article makes the following additional contributions:
\begin{enumerate}
    \item A revisited architectural design of the camera-conditioned monocular metric depth estimator network, which makes \ourmodel simpler, substantially more efficient in computation time and parameters, and at the same time more accurate than UniDepth. This design upgrade pertains to the simplification of the connections between the Camera Module and the Depth Module of the network, the more economic sinusoidal embedding of the pinhole-based dense camera representations fed to the Depth Module that we newly adopt, the inclusion of multi-resolution features and convolutional layers in our depth decoder, and the application of the geometric invariance loss solely on output-space features.
    \item A novel edge-guided scale-shift-invariant loss, which is computed from the predicted and ground-truth depth maps around geometric edges of the input, encourages \ourmodel to preserve the local structure of the depth map better, and thus enhances the sharpness of depth outputs substantially compared to UniDepth even on camera and scene domains which are unseen during training.
    \item An improved practical training strategy that presents the network with a greater diversity of input image shapes and resolutions within each mini-batch and hence with a larger range of intrinsic parameters of the assumed pinhole camera model, leading to increased robustness to the specific input distribution during inference.
    \item An additional, uncertainty-level output, which requires no additional supervisory signal during training yet allows to quantify confidence during inference reliably and thus enables downstream applications to geometric perception, \eg{} control, which require confidence-aware depth inputs.
\end{enumerate}
The methodological novelties introduced lead to improved performance, robustness, and efficiency of \ourmodel compared to UniDepth across a wide range of camera and scene domains.
This is demonstrated through an extensive set of comparisons to the latest state-of-the-art methods as well as ablation studies on 10 depth estimation benchmarks, both in the challenging zero-shot evaluation setting and in the practical supervised fine-tuning setting.
\ourmodel sets the overall \emph{new state of the art} in MMDE and ranks first among published methods in the competitive official public KITTI Depth Prediction Benchmark.}

\section{Related Work}
\label{sec:relwork}

\begin{figure*}[ht]
    \centering
    \includegraphics[width=1.0\linewidth]{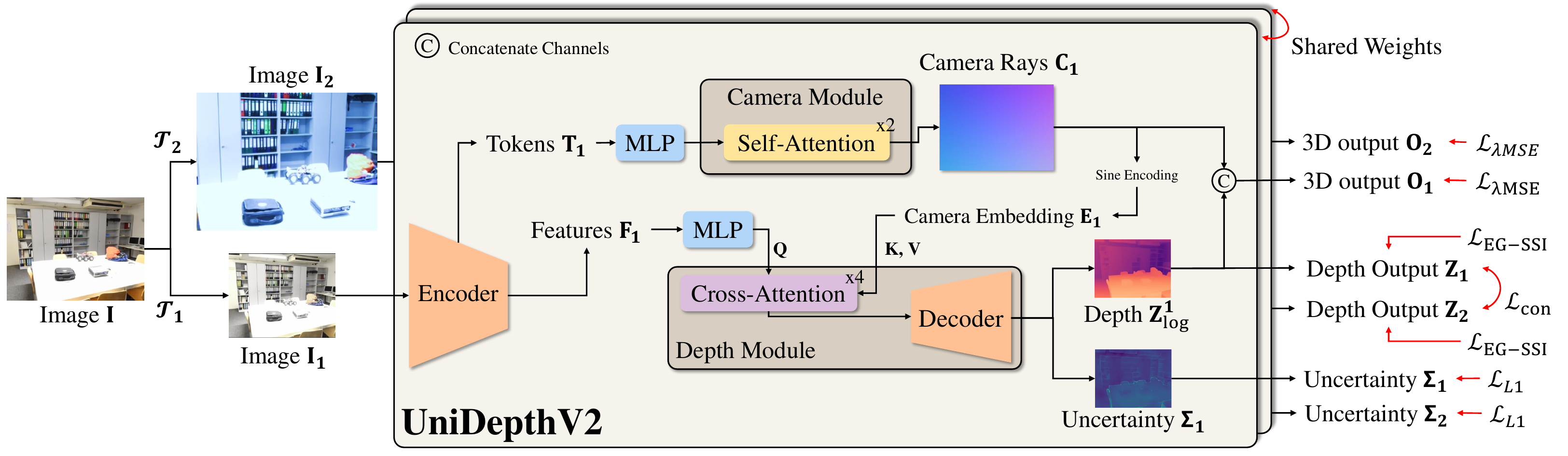}
    \caption{\textbf{Model Architecture.} \ourmodel utilizes solely the input image to generate the 3D output ($\mathbf{O}$). It bootstraps a dense camera prediction ($\mathbf{C}$) from the Camera Module, injecting prior knowledge on scene scale into the Depth Module via a cross-attention layer per resolution, with 4 layers in total. The camera representation corresponds to azimuth and elevation angles. The geometric invariance loss ($\mathcal{L}_{\mathrm{con}}$) enforces consistency between geometric camera-aware output tensors from different geometric augmentations ($\mathcal{T}_1$, $\mathcal{T}_2$). The depth output ($\mathbf{Z}_{\log}$) \blue{is obtained through an FPN-based decoder that gradually upsamples the feature maps and injects multi-resolution information}. The final output is the concatenation of the camera and depth tensors ($\mathbf{C} || \mathbf{Z}_{\log}$), creating two independent optimization spaces for $\mathcal{L}_{\lambda MSE}$. \blue{The depth output is supervised with the proposed Edge-guided Normalized L1-loss $\mathcal{L}_{EG-SSI}$. In addition, \ourmodel computes a prediction uncertainty ($\mathbf{\Sigma}$) which is supervised with an L1-loss on the error in log space between predicted and ground-truth depth.}}
    \label{fig:results:overview}
    \vspace{-1em}
\end{figure*}

\PAR{Metric and Scale-Agnostic Depth Estimation.}
It is crucial to distinguish Monocular Metric Depth Estimation (MMDE) from scale-agnostic, namely up-to-a-scale, monocular depth estimation.
MMDE SotA approaches typically confine training and testing to the same domain.
However, challenges arise, such as overfitting to the training scenario, leading to considerable performance drops in the presence of minor domain gaps,
often overlooked in benchmarks like NYU-Depthv2~\cite{silberman2012nyu} (NYU) and KITTI~\cite{Geiger2012kitti}.
On the other hand, scale-agnostic depth methods, pioneered by MiDaS~\cite{ranftl2020midas}, OmniData~\cite{eftekhar2021omnidata}, and LeReS~\cite{yin2021leres}, show robust generalization by training on extensive datasets.
The paradigm has been elevated to another level by repurposing depth-conditioned generative methods for RGB to RGB-conditioned depth generative methods~\cite{ke2024marigold} or large-scale semi-supervised pre-training as in the DepthAnything series~\cite{yang2024da1, yang2024da2}.
\bluereb{In particular, these two paradigms have been used extensively for other dense prediction tasks, such as segmentation~\cite{kirillov2023sam} and surface normal estimation~\cite{fu2024geowizard,he2025lotus}, and for other modalities, notably video~\cite{chen2025vda,hu2025depthcrafter}.}
The limitation of all these methods lies in the absence of a metric output, hindering practical usage in downstream applications.

\PAR{Monocular Metric Depth Estimation.}
The introduction of end-to-end trainable neural networks in MMDE, pioneered by \cite{Eigen2014}, marked a significant milestone, also introducing the optimization process through the Scale-Invariant log loss ($\mathrm{SI}_{\log}$).
Subsequent developments witnessed the emergence of advanced networks, ranging from convolution-based architectures~\cite{Fu2018Dorn, Laina2016, Liu2015, Patil2022p3depth} to transformer-based approaches~\cite{Yang2021, Bhat2020adabins, Yuan2022newcrf, piccinelli2023idisc, piccinelli2025velodepth}.
Despite impressive achievements on established benchmarks, MMDE models face challenges in zero-shot scenarios, revealing the need for robust generalization against appearance and geometry domain shifts.

\PAR{General Monocular Metric Depth Estimation.}
Recent efforts focus on developing MMDE models~\cite{bhat2023zoedepth, guizilini2023zerodepth, yin2023metric3d} for general depth prediction across diverse domains.
These models often leverage camera awareness, either by directly incorporating external camera parameters into computations~\cite{facil2019camconvs, guizilini2023zerodepth} or by normalizing the shape or output depth based on intrinsic properties, as seen in~\cite{Lee2019bts, Lopez2020mapillary, yin2023metric3d, hu2024metric3dv2}.
A new paradigm recently emerged~\cite{piccinelli2024unidepth, bochkovskii2024depthpro, piccinelli2025unik3d}, where the goal is to directly estimate the 3D scene from the input image \emph{without any} additional information other than the RGB input.
Our approach fits in the latter new paradigm, namely universal MMDE: we do not require any additional prior information at test time, such as access to camera information.

\bluereb{\PAR{Depth Estimation under Challenging Conditions.}
A parallel line of work targets robustness when image formation departs from the Lambertian, well-lit assumption.
Diffusion-augmented training and guidance improve performance in rare or adverse conditions (\eg night, fog, rain, low light)~\cite{tosi2024diffusion}, while dedicated training/evaluation protocols formalize the ``challenging conditions'' setting and report consistent gains across benchmarks~\cite{gasperini2023robustmde}.
For non-Lambertian materials, methods tailored to transparent and mirror surfaces mitigate severe failure modes~\cite{costanzino2023transparent,zamaramirez2024booster}.
Our work is complementary: we pursue universal MMDE, aiming for a single camera-free model that generalizes across such conditions through large-scale training.}

\section{\ourmodel}
\label{sec:method}

\blue{Most of} the SotA MMDE methods typically assume access to the camera intrinsics, thus blurring the line between pure depth estimation and actual 3D estimation.
In contrast, \ourmodel aims to create a universal MMDE model deployable in diverse scenarios without relying on any other external information, such as camera intrinsics, thus leading to 3D-space estimation by design.
However, attempting to directly predict 3D points from a single image without a proper internal representation neglects geometric prior knowledge, \ie perspective geometry, burdening the learning process with re-learning laws of perspective projection from data.

\cref{ssec:method:spherical} introduces a pseudo-spherical representation of the output space to inherently disentangle camera rays' angles from depth.
In addition, our preliminary studies indicate that depth prediction benefits from prior information on the acquisition sensor, leading to the introduction of a self-prompting camera operation in \cref{ssec:method:camera_module}.
Further disentanglement at the level of \blue{depth prediction} is achieved through a geometric invariance loss, outlined in \cref{ssec:method:consistency}.
This loss ensures \blue{depth predictions} remain invariant when conditioned on the bootstrapped camera predictions, promoting robust camera-aware depth predictions.
\blue{Furthermore, the spatial resolution is enhanced via an edge-guided normalized loss on the depth prediction that forces the network to learn both sharp transitions in depth values and flat surfaces.}
The overall architecture and the resulting optimization induced by the combination of design choices are detailed in \cref{ssec:method:design}.

\subsection{3D Representation}
\label{ssec:method:spherical}

The general-purpose nature of our MMDE method requires inferring both depth and camera intrinsics to make 3D predictions based only on imagery observations.
We design the 3D output space presenting a natural disentanglement of the two sub-tasks, namely depth estimation and camera calibration.
In particular, we exploit the pseudo-spherical representation where the basis is defined by azimuth, elevation, and log-depth, \ie ($\theta$,$\phi$,$z_{\log}$), in contrast to the Cartesian representation ($x$,$y$,$z$).
The strength of the proposed pseudo-spherical representation lies in the decoupling of camera ($\theta$,$\phi$) and depth ($z_{\log}$) components, ensuring their orthogonality by design, in contrast to the entanglement present in Cartesian representation.

It is worth highlighting that in this output space, the non-parametric dense representation of the camera is mathematically represented as a tensor $\mathbf{C} \in \mathbb{R}^{H \times W \times 2}$, where $H$ and $W$ are the height and width of the input image and the last dimension corresponds to azimuth and elevation values.
While in the typical Cartesian space, the backprojection involves the multiplication of homogeneous camera rays and depth, the backprojection operation in the proposed representation space accounts for the concatenation of camera and depth representations.
The pencil of rays are defined as $(\mathbf{r}_1, \mathbf{r}_2, \mathbf{r}_3) = \mathbf{K}^{-1} [\mathbf{u}, \mathbf{v}, \mathbf{1}]^T$, where $\mathbf{K}$ is the calibration matrix, $\mathbf{u}$ and $\mathbf{v}$ are pixel positions in pixel coordinates, and $\mathbf{1}$ is a vector of ones. Therefore, the homogeneous camera rays $(\mathbf{r}_x, \mathbf{r}_y)$ correspond to $(\frac{\mathbf{r}_1}{\mathbf{r}_3}, \frac{\mathbf{r}_2}{\mathbf{r}_3})$.
\blue{Moreover, this dense camera representation can be embedded via a standard Sine encoding, where the total amount of harmonics is 64 per homogeneous ray dimension, namely 128 channels in total.}

\subsection{Self-Promptable Camera}
\label{ssec:method:camera_module}

The camera module plays a crucial role in the final 3D predictions since its angular dense output accounts for two dimensions of the output space, namely azimuth and elevation.
Most importantly, these embeddings prompt the depth module to ensure a bootstrapped prior knowledge of the input scene's global depth scale.
The prompting is fundamental to avoid mode collapse in the scene scale and to alleviate the depth module from the burden of predicting depth from scratch as the scale is already modeled by camera output.

Nonetheless, the internal representation of the camera module is based on a pinhole parameterization, namely via focal length ($f_x$, $f_y$) and principal point ($c_x$, $c_y$).
The four tokens conceptually corresponding to the intrinsics are then projected to scalar values, \ie, $\Delta f_x$, $\Delta f_y$, $\Delta c_x$, $\Delta c_y$.
However, they do not directly represent the camera parameters, but the multiplicative residuals to a pinhole camera initialization, namely $\frac{H}{2}$ for y-components and $\frac{W}{2}$ for x-components, leading to $f_x = \frac{\Delta f_x W}{2}$, $f_y = \frac{\Delta f_y H}{2}$, $c_x = \frac{\Delta c_x W}{2}$, $c_y = \frac{\Delta c_y H}{2}$, leading to invariance towards input image sizes.

Subsequently, a backprojection operation based on the intrinsic parameters is applied to every pixel coordinate to produce the corresponding rays.
The rays are normalized and thus represent vectors on a unit sphere.
The critical step involves extracting azimuth and elevation from the backprojected rays, effectively creating a ``dense'' angular camera representation.
This dense representation undergoes \blue{Sine encoding} to produce the embeddings $\mathbf{E}$.
The embedded representations are then seamlessly passed to the depth module as a prompt, where they play a vital role as a conditioning factor.
The conditioning is enforced via a cross-attention layer between \blue{the projected encoder feature maps $\{\mathcal{F}_i\}^{4}_{i=1}$, with $\mathbf{F}_i \in \mathbb{R}^{h \times w \times C}$} and the camera embeddings $\mathbf{E}$ where $(h,w)=(H/14, W/14)$. The camera-prompted depth features $\mathbf{F}_i|\mathbf{E} \in \mathbb{R}^{h \times w \times C}$ are defined as 
\begin{equation}
    \mathbf{F}_i|\mathbf{E} = \mathrm{MLP}(\mathrm{CA}(\mathbf{F}_i, \mathbf{E})),
\end{equation}
where $\mathrm{CA}$ is a cross-attention block and $\mathrm{MLP}$ is a MultiLayer Perceptron with one $4C$-channel hidden layer.

\subsection{Geometric Invariance Loss}
\label{ssec:method:consistency}

The spatial locations from the same scene captured by different cameras should correspond when the depth module is conditioned on the specific camera.
To this end, we propose a geometric invariance loss to enforce the consistency of camera-prompted depth features of the same scene from different acquisition sensors.
In particular, consistency is enforced on features extracted from identical 3D locations.

For each image, we perform $N$ distinct geometrical augmentations, denoted as $\{\mathcal{T}_i\}_{i=1}^N$, with $N=2$ in our experiments. 
This operation involves sampling a rescaling factor \blue{$r \sim 2^{\mathcal{U}_{[-2, 2]}}$} and a relative translation $t \sim \mathcal{U}_{[-0.1, 0.1]}$, then cropping it to the current step randomly selected input shape.
This is analogous to sampling a pair of images from the same scene and extrinsic parameters but captured by different cameras.
Let $\mathbf{C}_i$ and $\mathbf{Z}_i$ describe the predicted camera representation and \blue{camera-aware depth output}, respectively, corresponding to augmentation $\mathcal{T}_i$. 
It is evident that the camera representations differ when two diverse geometric augmentations are applied, i.e., $\mathbf{C}_i \neq \mathbf{C}_j$ if $\mathcal{T}_i \neq \mathcal{T}_j$. 
Therefore, the geometric invariance loss can be expressed as
\begin{equation}
    \mathcal{L}_{\mathrm{con}}(\mathbf{Z}_1, \mathbf{Z}_2) =\\
    \lVert \mathcal{T}_2 \circ \mathcal{T}^{-1}_1 \circ (\mathbf{Z}_1) - \mathrm{sg}(\mathbf{Z}_2) \rVert_1,
\label{eqn:method:selfconst}
\end{equation}
where $\mathbf{Z}_i$ represents the depth output after being conditioned by camera prompt $\mathbf{E}_i$, as outlined in \cref{ssec:method:camera_module}, \blue{and decoded}; $\mathrm{sg}(\cdot)$ corresponds to the stop-gradient detach operation needed to exploit $\mathbf{Z}_2$ as pseudo ground truth (GT).
The bidirectional loss can be computed as: $\frac{1}{2} (\mathcal{L}_{\mathrm{con}}(\mathbf{Z}_1, \mathbf{Z}_2) + \mathcal{L}_{\mathrm{con}}(\mathbf{Z}_2, \mathbf{Z}_1))$.
It is necessary to apply the geometric invariance loss \blue{on the components that are camera-aware, such as the output depth map.}
Otherwise, the loss would enforce consistency across features that carry camera information purposely different.

\subsection{Edge-Guided Normalized Loss}
\label{ssec:method:egssi}

Modern depth estimation methods must balance global scene understanding with local geometric precision.
While UniDepth excels at the former, it lacks accuracy in local, fine-grained details of the geometry of the depicted scenes. To address this, \ourmodel involves a novel loss function, named Edge-Guided Scale-Shift Invariant Loss ($\mathcal{L}_{\mathrm{EG-SSI}}$), which is explicitly designed to enhance local precision.
This loss is computed over image patches extracted from regions where the RGB spatial gradient ranks in the top 5\%-quantile, capturing high-contrast areas likely to contain depth discontinuities.
Patch sizes are randomly sampled between 4\% and 8\% of the input image's smallest dimension.
By concentrating on these visually salient regions, our model learns to distinguish between genuine geometric discontinuities and misleading high-frequency textures that do not correspond to actual depth changes.
\bluereb{The loss oversamples high RGB-gradient regions so that thin, high-frequency boundaries are well represented.
We do not use superpixels (\eg SLIC~\cite{achanta2012slic}) because their emphasis on region uniformity suppresses thin structures and creates a non-maximum-suppression-like effect that harms boundary detail.}
For instance, structured patterns such as checkerboard textures or repetitive details on flat surfaces can falsely suggest depth variations, leading to hallucinated discontinuities.

Our approach discourages such errors by enforcing local consistency between the predicted and ground-truth depth.
At each selected patch location, we apply a local normalization step where both the predicted depth and ground-truth depth are independently aligned in scale and shift based on the patch’s statistics.
This ensures that the loss directly measures shape consistency rather than absolute depth values, making it robust to variations in depth scale across different scenes.
Specifically, our loss function is formulated as:
\begin{equation}
    \mathcal{L}_{\mathrm{EG-SSI}}(\mathbf{D}, \mathbf{D^*}, \Omega) = \sum_{\omega \in \Omega} \left|| \mathcal{N}_\omega (\mathbf{D}_{\omega}) - \mathcal{N}_\omega(\mathbf{D}^*_{\omega})\right||_1,
\label{eqn:method:egssi}
\end{equation}
where $\mathbf{D}$ and $\mathbf{D}^*$ are the predicted and ground-truth inverse depth, $\Omega$ is the set of extracted RGB patches, and $\mathbf{D}_{\omega}$ represents depth values within patch $\omega$.
The function $\mathcal{N}_\omega(\cdot)$ denotes the standardization operation via subtracting the median and dividing by the mean absolute deviation (MAD) over the patch $\omega$.
A key advantage of this formulation is that it penalizes two distinct failure cases: (i) regions where the model ignores strong chromatic cues, failing to capture a true depth discontinuity, and (ii) regions where the model incorrectly exploits changes solely in appearance, hallucinating depth discontinuities that do not correspond to actual geometric edges.
Since random patch extraction is computationally inefficient in standard ML frameworks such as PyTorch, we implement a custom CUDA kernel, accelerating loss computation by 20x.

\begin{figure*}[t]
    \renewcommand{\arraystretch}{3}
    \centering
    \small
    \begin{tabular}{cc|cccc|cc}
        \multirow{1}{*}[4.8em]{\rotatebox[origin=c]{90}{IBims-1}}
        & \includegraphics[width=0.14\linewidth]{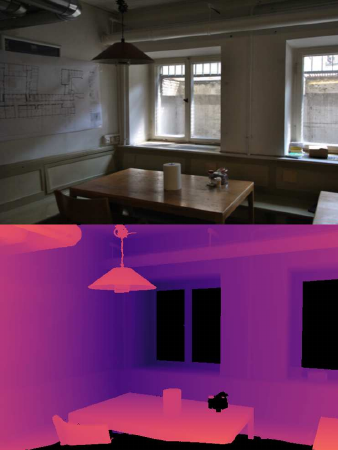}
        & \includegraphics[width=0.14\linewidth]{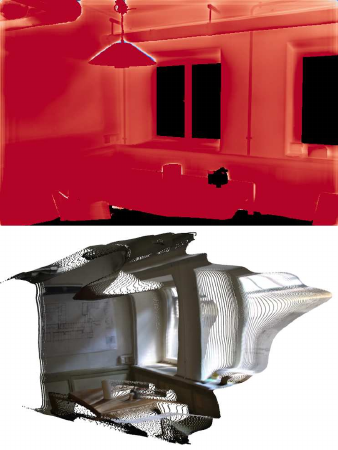}
        & \includegraphics[width=0.14\linewidth]{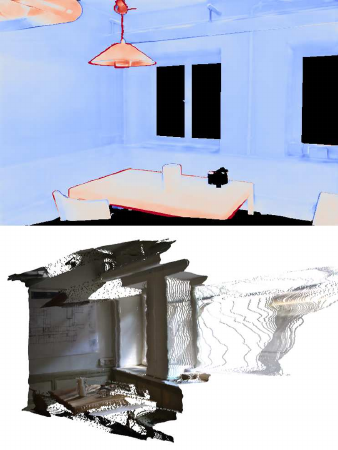}
        & \includegraphics[width=0.14\linewidth]{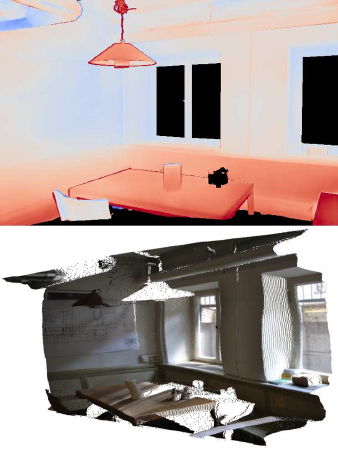}
        & \includegraphics[width=0.14\linewidth]{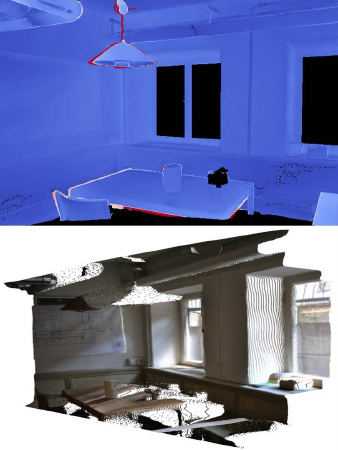}
        & \includegraphics[width=0.045\linewidth]{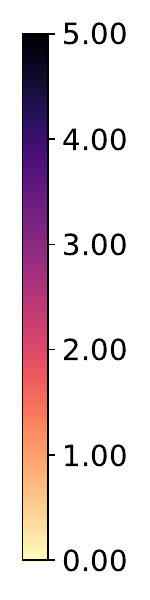}
        & \includegraphics[width=0.045\linewidth]{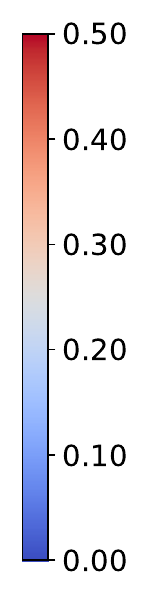}\\

        \multirow{1}{*}[5.5em]{\rotatebox[origin=c]{90}{TUM-RGBD}}
        & \includegraphics[width=0.14\linewidth]{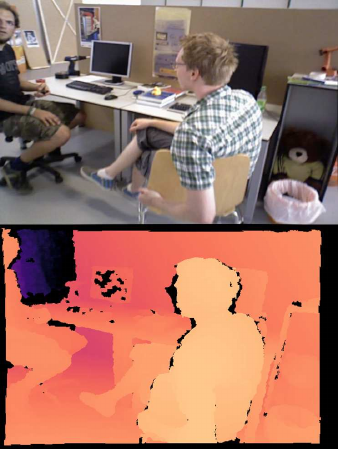}
        & \includegraphics[width=0.14\linewidth]{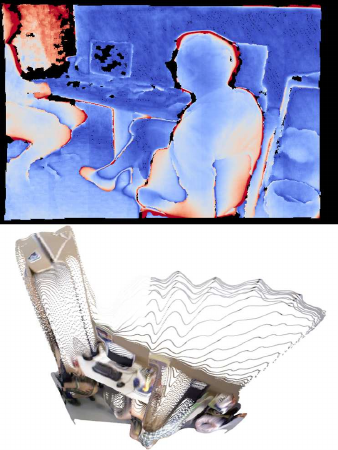}
        & \includegraphics[width=0.14\linewidth]{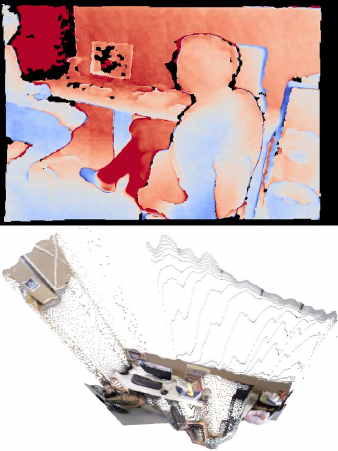}
        & \includegraphics[width=0.14\linewidth]{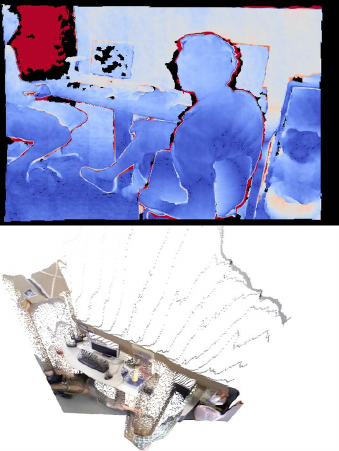}
        & \includegraphics[width=0.14\linewidth]{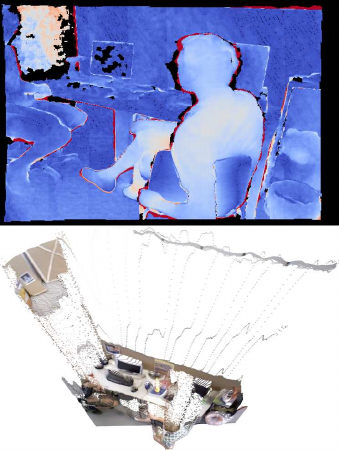}
        & \includegraphics[width=0.045\linewidth]{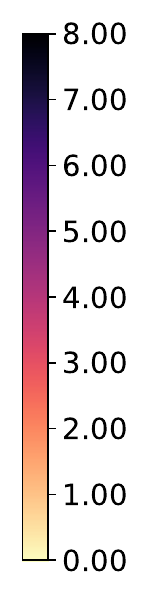}
        & \includegraphics[width=0.045\linewidth]{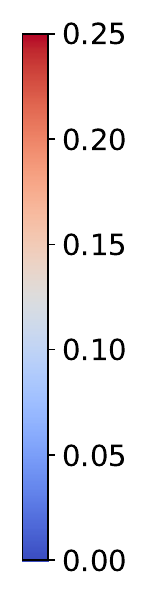} \\

        \multirow{1}{*}[4.8em]{\rotatebox[origin=c]{90}{DDAD}}
        & \includegraphics[width=0.14\linewidth]{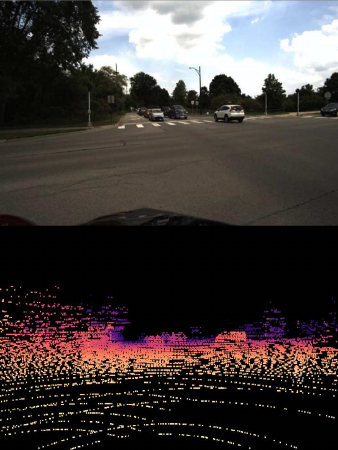}
        & \includegraphics[width=0.14\linewidth]{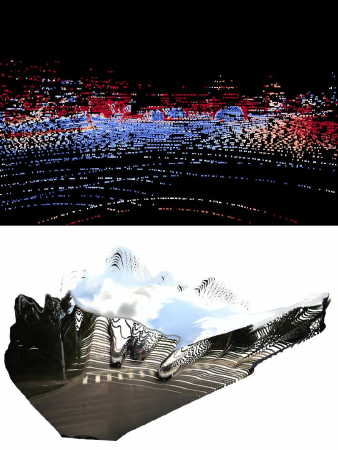}
        & \includegraphics[width=0.14\linewidth]{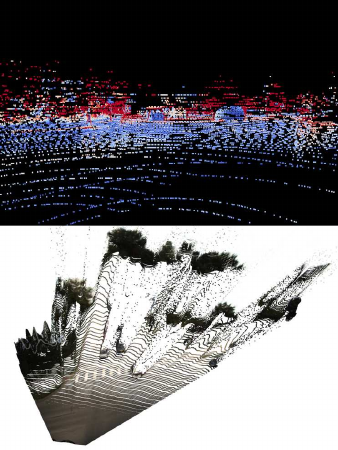}
        & \includegraphics[width=0.14\linewidth]{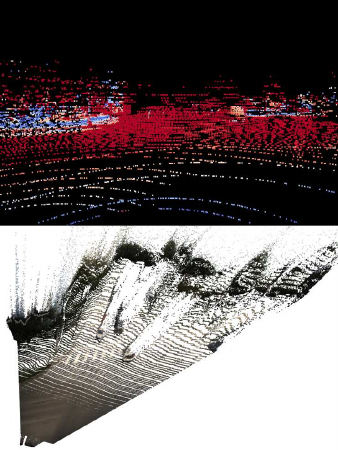}
        & \includegraphics[width=0.14\linewidth]{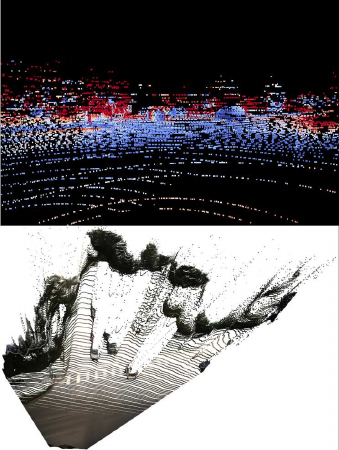}
        & \includegraphics[width=0.045\linewidth]{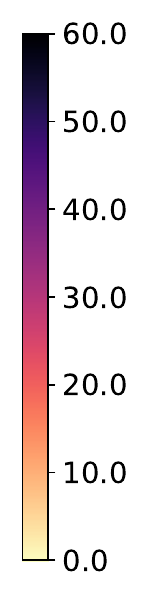}
        & \includegraphics[width=0.045\linewidth]{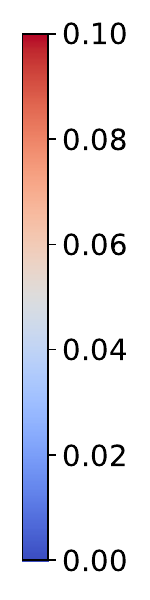} \\

        \multirow{1}{*}[4.8em]{\rotatebox[origin=c]{90}{Sintel}}
        & \includegraphics[width=0.14\linewidth]{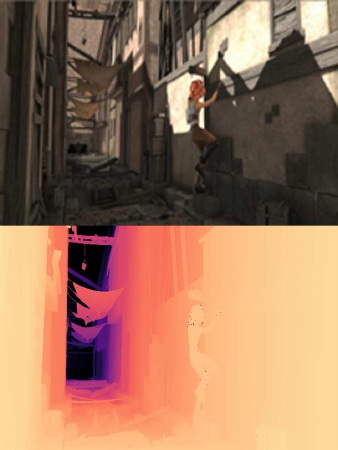}
        & \includegraphics[width=0.14\linewidth]{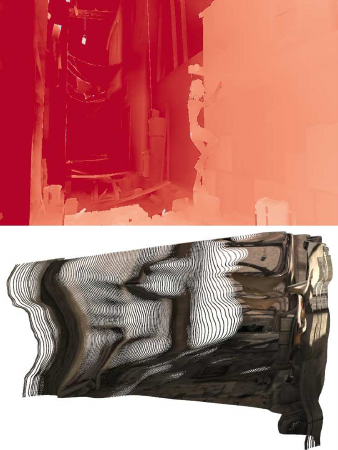}
        & \includegraphics[width=0.14\linewidth]{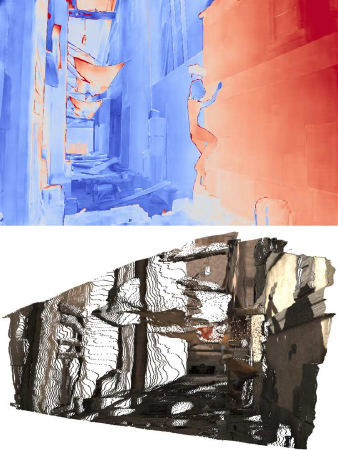}
        & \includegraphics[width=0.14\linewidth]{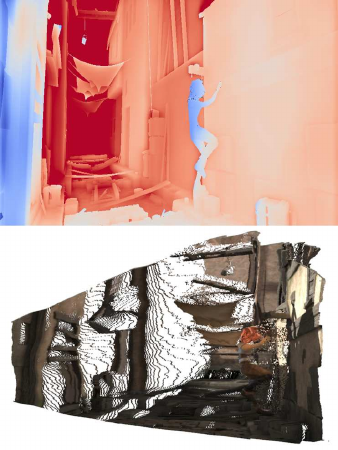}
        & \includegraphics[width=0.14\linewidth]{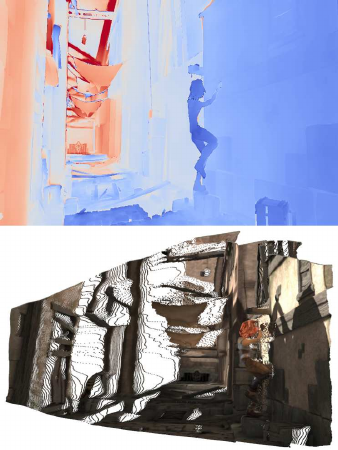}
        & \includegraphics[width=0.045\linewidth]{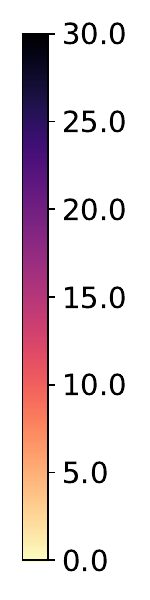}
        & \includegraphics[width=0.045\linewidth]{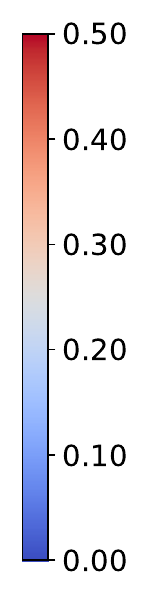}\\
        
        & RGB \& GT & UniDepth~\cite{piccinelli2024unidepth} & Metric3Dv2\textsuperscript{\dag}~\cite{hu2024metric3dv2} & DepthPro~\cite{bochkovskii2024depthpro} & \ourmodel & Depth & $\mathrm{A.Rel}$ \\
    \end{tabular}
    \caption{\textbf{Zero-shot qualitative results.} Each pair of consecutive rows corresponds to one test sample. Each odd row shows the input RGB image and the 2D error map color-coded with \textit{coolwarm} based on the absolute relative error. Each even row shows GT depth and the predicted point cloud. The last column represents the specific colormap ranges for depth and error. (\dag): DDAD domain in the training set.}
    \label{fig:results:main_vis}
    \vspace{-1em}
\end{figure*}

\subsection{Network Design}
\label{ssec:method:design}

\PAR{Architecture.} Our network, described in \cref{fig:results:overview}, comprises an Encoder Backbone, a Camera Module, and a Depth Module.
The encoder is ViT-based~\cite{Dosovitskiy2020VIT}, producing features at four different ``scales'', \blue{\ie $\{\mathbf{F}_i\}^{4}_{i=1}$, with $\mathbf{F}_i \in \mathbb{R}^{h \times w \times C}$, where $(h,w) = (\frac{H}{14}, \frac{W}{14})$.}

The four Camera Module parameters are initialized as class tokens present in ViT-style backbones.
\blue{After this initialization, they are (i) processed via 2 layers of self-attention to obtain the corresponding pinhole parameters which are used to produce} the final dense representation $\mathbf{C}$ as detailed in \cref{ssec:method:camera_module}, and (ii) further embedded to $\mathbf{E}$ \blue{via a Sine encoding.}

\blue{The Depth Module is fed with the four feature maps $\{\mathbf{F}_i\}^{4}_{i=1}$ from the encoder.
Each feature map $\mathbf{F}_i$ is conditioned on the camera prompts $\mathbf{E}$ to obtain $\mathbf{D|E}$ as described in \cref{ssec:method:camera_module} with a different cross-attention layer.
The four feature maps are then processed with an FPN-style decoder where the ``lateral'' convolution is transposed convolution to match the ViT resolution to the resolution of the different layers of the FPN.}
The log-depth prediction $\mathbf{Z}_{\log} \in \mathbb{R}^{H \times W \times 1}$ corresponds to the \blue{last FPN feature map which is upsampled to the original input shape and processed with two convolutional layers.}
The final 3D output $\mathbf{O} \in \mathbb{R}^{H \times W \times 3}$ is the concatenation of predicted rays and depth, $\mathbf{O} = \mathbf{C} || \mathbf{Z}$, with $\mathbf{Z}$ as element-wise exponentiation of $\mathbf{Z}_{\log}$.
\bluereb{Owing to the architecture’s modularity, the ray bundle $\mathbf{C}$ need not come from \ourmodel's Camera Module: at inference, it can be injected from any camera model with known parameters and a specified unprojection operator.}

\PAR{Optimization.} The optimization process is guided by a re-formulation of the Mean Squared Error (MSE) loss in the final 3D output space ($\theta$,$\phi$,$z_{\log}$) from \cref{ssec:method:spherical} as:
\vspace{-4pt}
\begin{equation}
    \vspace{-4pt}
    \begin{split}
        \mathcal{L}_{\lambda\mathrm{MSE}}(\bm{\varepsilon}) = \|\mathbb{V}[\bm{\varepsilon}]\|_1 + \bm{\lambda}^T(\mathbb{E}[\bm{\varepsilon}]\odot\mathbb{E}[\bm{\varepsilon}]),
    \end{split}
    \label{eqn:method:mse}
\end{equation}
where $\bm{\varepsilon} = \hat{\mathbf{o}} - \mathbf{o}^* \in \mathbb{R}^3$, $\hat{\mathbf{o}}=(\hat{\theta},\hat{\phi},\hat{z}_{\log})$ is the predicted 3D output, $\mathbf{o}^*=(\theta^*,\phi^*,z_{\log}^*)$ is the GT 3D value, and $\bm{\lambda} = (\lambda_{\theta},\lambda_{\phi},\lambda_z) \in \mathbb{R}^3$ is a vector of weights for each dimension of the output.
$\mathbb{V}[\bm{\varepsilon}]$ and $\mathbb{E}[\bm{\varepsilon}]$ are computed as the vectors of empirical variances and means for each of the three output dimensions over all pixels, \ie $\{\bm{\varepsilon}^{i}\}_{i=1}^{N}$. 
Note that if $\lambda_d=1$ for a dimension $d$, the loss represents the standard MSE loss for that dimension. If $\lambda_d<1$, a scale-invariant loss term is added to that dimension if it is expressed in log space, \eg{}for the depth dimension $z_{\log}$, or a shift-invariant loss term is added if that output is expressed in linear space.
In particular, if only the last output dimension is considered, \ie{}the one corresponding to depth, and $\lambda_z=0.15$ is utilized, the corresponding loss is the standard $\mathrm{SI}_{\log}$.
In our experiments, we set $\lambda_{\theta}=\lambda_{\phi}=1$ and $\lambda_z=0.15$.
In addition, we extended the optimization with the supervision for the uncertainty prediction \bluereb{$\Sigma$}, defined as an L1 loss between the predicted uncertainty and the detached error in log space between predicted depth ($\mathbf{Z}_{\log}$) and GT depth ($\mathbf{Z}^{*}_{\log}$). More formally,
\begin{equation} \label{eqn:loss:uncertainty}
\mathcal{L}_{\mathrm{L1}} = \lVert \mathbf{\Sigma} - \mathrm{sg}(| \mathbf{Z}_{\log} - \mathbf{Z}^{*}_{\log} |) \rVert_{1},
\end{equation}
\blue{with $\mathrm{sg(\cdot)}$ referring to the stop gradient operation.}
Therefore, the final optimization loss is defined as
\begin{equation}
    \begin{split}
        \mathcal{L} = \mathcal{L}_{\lambda\mathrm{MSE}} + \alpha \mathcal{L}_{\mathrm{con}} + \beta \mathcal{L}_{\mathrm{EG-SSI}} + \gamma \mathcal{L}_{\mathrm{L1}},\\ \text{ with } (\alpha,\beta,\gamma) =(0.1,1.0,0.1).
    \end{split}
    \label{eqn:method:loss}
\end{equation}

The loss defined here serves as a motivation for the designed output representation.
Specifically, employing a Cartesian representation and applying the loss directly to the output space would result in backpropagation through ($x$, $y$), and $z_{\log}$ errors.
However, $x$ and $y$ components are derived as $r_x \cdot z$ and $r_y \cdot z$ as detailed in \cref{ssec:method:spherical}.
Consequently, the gradients of camera components, expressed by ($r_x$, $r_y$), and of depth become intertwined, leading to suboptimal optimization as discussed in~\cref{ssec:experiments:ablations}.
\blue{Depth estimators often entangle image shape with scene scale by implicitly encoding aspects of the camera parameters within the image dimensions~\cite{yin2023metric3d}.
This reliance on fixed input shapes can limit their ability to generalize across different image resolutions and aspect ratios.
In contrast, \ourmodel is designed to be robust to variations in image shape, ensuring that the predicted scene geometry and camera FoV remain consistent regardless of input resolution.
This flexibility allows the model to adapt to different computational constraints, striking a balance between finer detail and processing speed while maintaining global scene accuracy.
To achieve this robustness, we train on dynamically varying image shapes and resolutions, ensuring that the model learns to infer depth consistently across a wide range of input conditions.
Specifically, we sample images with variable pixel counts between 0.2MP and 0.6MP, allowing the model to operate effectively across diverse resolutions without being biased toward a single fixed input size.}

\section{Experiments}
\label{sec:experiments}

\begin{table*}[t]
    \centering
    \caption{\textbf{Results for Indoor Domains.} All methods are tested in a zero-shot fashion. Missing values (\textcolor{gray}{-}) indicate the model's inability to produce the respective output. \dag: Requires ground-truth (GT) camera for 3D reconstruction. \ddag: Requires GT camera for 2D depth map inference.}
    \label{tab:results:indoor}
    \vspace{-1em}
    \resizebox{\linewidth}{!}{%
    \begin{tabular}{l|cccc|cccc|cccc}
    \toprule
    \multirow{2}{*}{\textbf{Method}} 
      & \multicolumn{4}{c|}{SUNRGBD} 
      & \multicolumn{4}{c|}{IBims-1} 
      & \multicolumn{4}{c}{TUM-RGBD} \\
      & $\mathrm{\delta_1}\uparrow$ & $\mathrm{ARel}\downarrow$ & $F_1\uparrow$ & $\rho\uparrow$
      & $\mathrm{\delta_1}\uparrow$ & $\mathrm{ARel}\downarrow$ & $F_1\uparrow$ & $\rho\uparrow$
      & $\mathrm{\delta_1}\uparrow$ & $\mathrm{ARel}\downarrow$ & $F_1\uparrow$ & $\rho\uparrow$ \\
    \midrule
    Metric3D\textsuperscript{\dag \ddag}~\cite{yin2023metric3d}
      & $1.9$ & \bluereb{$48.7$} & \textcolor{gray}{-} & \textcolor{gray}{-}
      & $75.1$ & \bluereb{$19.3$} & \textcolor{gray}{-} & \textcolor{gray}{-}
      & $7.7$ & \bluereb{$61.1$} & \textcolor{gray}{-} & \textcolor{gray}{-} \\
    Metric3Dv2\textsuperscript{\dag \ddag}~\cite{hu2024metric3dv2}
      & $81.2$ & \bluereb{$13.3$} & \textcolor{gray}{-} & \textcolor{gray}{-}
      & $68.4$ & \bluereb{$20.7$} & \textcolor{gray}{-} & \textcolor{gray}{-}
      & $63.0$ & \bluereb{$\best{9.3}$} & \textcolor{gray}{-} & \textcolor{gray}{-} \\
    ZoeDepth\textsuperscript{\dag}~\cite{bhat2023zoedepth}
      & $80.9$ & \bluereb{$13.6$} & \textcolor{gray}{-} & \textcolor{gray}{-}
      & $49.8$ & \bluereb{$21.5$} & \textcolor{gray}{-} & \textcolor{gray}{-}
      & $55.6$ & \bluereb{$33.6$} & \textcolor{gray}{-} & \textcolor{gray}{-} \\
    UniDepth~\cite{piccinelli2024unidepth}
      & $94.3$ & \bluereb{$10.4$} & $78.6$ & $85.8$
      & $15.7$ & \bluereb{$41.0$} & $30.3$ & $\best{76.6}$
      & $72.3$ & \bluereb{$\scnd{17.2}$} & $54.8$ & $86.8$ \\
    MASt3R~\cite{leroy2024master}
      & $80.1$ & \bluereb{$14.4$} & $71.5$ & $\scnd{92.0}$
      & $61.0$ & \bluereb{$19.7$} & $55.7$ & $76.0$
      & $52.4$ & \bluereb{$27.9$} & $44.1$ & $\scnd{93.7}$ \\
    DepthPro~\cite{bochkovskii2024depthpro}
      & $83.1$ & \bluereb{$13.3$} & $71.1$ & $89.3$
      & $82.3$ & \bluereb{$17.0$} & $62.8$ & $75.9$
      & $56.9$ & \bluereb{$19.9$} & $48.1$ & $\best{96.5}$ \\
    \midrule
    \ourmodel-Small
      & $90.8$ & \bluereb{$10.5$} & $74.2$ & $87.7$
      & $86.6$ & \bluereb{$13.5$} & $62.4$ & $67.5$
      & $69.0$ & \bluereb{$23.6$} & $50.6$ & $86.1$ \\
    \ourmodel-Base
      & $\scnd{94.4}$ & \bluereb{$\scnd{8.4}$} & $\scnd{79.9}$ & $91.1$
      & $\scnd{89.7}$ & \bluereb{$\scnd{11.1}$} & $\scnd{68.5}$ & $\scnd{76.5}$
      & $\scnd{77.5}$ & \bluereb{$20.7$} & $\scnd{57.3}$ & $89.4$ \\
    \ourmodel-Large
      & $\best{96.4}$ & \bluereb{$\best{6.8}$} & $\best{84.6}$ & $\best{93.4}$
      & $\best{94.5}$ & \bluereb{$\best{7.8}$} & $\best{70.9}$ & $74.1$
      & $\best{90.5}$ & \bluereb{$22.1$} & $\best{62.9}$ & $89.6$ \\
    \bottomrule
    \end{tabular}%
    }
\end{table*}

\begin{table*}[t]
    \centering
    \caption{\textbf{Results for Outdoor Domains.} All methods are tested in a zero-shot fashion. Missing values (\textcolor{gray}{-}) indicate the model's inability to produce the respective output. \dag: Requires ground-truth (GT) camera for 3D reconstruction. \ddag: Requires GT camera for 2D depth map inference. \bluereb{For DDAD, Metric3D and Metric3Dv2 are excluded from evaluation as they are trained on it.}}
    \label{tab:results:outdoor}
    \vspace{-1em}
    \resizebox{\linewidth}{!}{%
    \begin{tabular}{l|cccc|cccc|cccc|cccc}
    \toprule
    \multirow{2}{*}{\textbf{Method}}  & \multicolumn{4}{c|}{ETH3D} & \multicolumn{4}{c|}{Sintel} & \multicolumn{4}{c|}{DDAD} & \multicolumn{4}{c}{NuScenes} \\
     & $\mathrm{\delta_1}\uparrow$ & $\mathrm{ARel}\downarrow$ & $\mathrm{F_1}\uparrow$ & $\mathrm{\rho}\uparrow$
     & $\mathrm{\delta_1}\uparrow$ & $\mathrm{ARel}\downarrow$ & $\mathrm{F_1}\uparrow$ & $\mathrm{\rho}\uparrow$
     & $\mathrm{\delta_1}\uparrow$ & $\mathrm{ARel}\downarrow$ & $\mathrm{F_1}\uparrow$ & $\mathrm{\rho}\uparrow$
     & $\mathrm{\delta_1}\uparrow$ & $\mathrm{ARel}\downarrow$ & $\mathrm{F_1}\uparrow$ & $\mathrm{\rho}\uparrow$ \\
    \midrule
    Metric3D\textsuperscript{\dag \ddag}~\cite{yin2023metric3d}
      & $19.7$ & \bluereb{$136.8$} & \textcolor{gray}{-} & \textcolor{gray}{-}
      & $1.4$  & \bluereb{$105.5$} & \textcolor{gray}{-} & \textcolor{gray}{-}
      & \textcolor{gray}{n/a} & \bluereb{\textcolor{gray}{n/a}} & \textcolor{gray}{n/a} & \textcolor{gray}{n/a}
      & $75.4$ & \bluereb{$23.7$} & \textcolor{gray}{-} & \textcolor{gray}{-} \\
    Metric3Dv2\textsuperscript{\dag \ddag}~\cite{hu2024metric3dv2}
      & $\best{90.0}$ & \bluereb{$\best{12.7}$} & \textcolor{gray}{-} & \textcolor{gray}{-}
      & $\best{34.5}$ & \bluereb{$\best{43.9}$} & \textcolor{gray}{-} & \textcolor{gray}{-}
      & \textcolor{gray}{n/a} & \bluereb{\textcolor{gray}{n/a}} & \textcolor{gray}{n/a} & \textcolor{gray}{n/a}
      & $84.1$ & \bluereb{$23.6$} & \textcolor{gray}{-} & \textcolor{gray}{-} \\
    ZoeDepth\textsuperscript{\dag}~\cite{bhat2023zoedepth}
      & $33.8$ & \bluereb{$54.7$} & \textcolor{gray}{-} & \textcolor{gray}{-}
      & $5.6$  & \bluereb{$95.9$} & \textcolor{gray}{-} & \textcolor{gray}{-}
      & $27.9$ & \bluereb{$50.9$} & \textcolor{gray}{-} & \textcolor{gray}{-}
      & $33.8$ & \bluereb{$42.0$} & \textcolor{gray}{-} & \textcolor{gray}{-} \\
    UniDepth~\cite{piccinelli2024unidepth}
      & $18.5$ & \bluereb{$53.3$} & $27.6$ & $42.6$
      & $13.2$ & \bluereb{$124.6$} & $40.2$ & $65.6$
      & $\scnd{85.8}$ & \bluereb{$\best{12.5}$} & $\scnd{72.8}$ & $\best{98.1}$
      & $84.6$ & \bluereb{$\best{12.7}$} & $\scnd{64.4}$ & $\best{97.7}$ \\
    MASt3R~\cite{leroy2024master}
      & $21.4$ & \bluereb{$45.3$} & $28.4$ & $\scnd{92.2}$
      & $17.2$ & \bluereb{$71.6$} & $41.5$ & $72.2$
      & $4.3$  & \bluereb{$56.7$} & $22.1$ & $74.6$
      & $2.7$  & \bluereb{$65.6$} & $13.6$ & $78.3$ \\
    DepthPro~\cite{bochkovskii2024depthpro}
      & $39.7$ & \bluereb{$65.2$} & $41.2$ & $77.4$
      & $26.2$ & \bluereb{$133.4$} & $49.7$ & $75.2$
      & $29.9$ & \bluereb{$37.3$} & $42.1$ & $83.0$
      & $56.6$ & \bluereb{$28.7$} & $46.5$ & $79.1$ \\
    \midrule
    \ourmodel-Small
      & $64.6$ & \bluereb{$113.2$} & $44.3$ & $78.4$
      & $14.6$ & \bluereb{$107.7$} & $37.1$ & $73.5$
      & $83.3$ & \bluereb{$16.4$} & $68.5$ & $94.7$
      & $82.1$ & \bluereb{$18.3$} & $59.7$ & $96.2$ \\
    \ourmodel-Base
      & $75.4$ & \bluereb{$70.1$} & $\scnd{53.5}$ & $91.4$
      & $31.9$ & \bluereb{$\scnd{50.1}$} & $\best{51.8}$ & $\scnd{75.9}$
      & $\scnd{86.8}$ & \bluereb{$14.2$} & $71.4$ & $96.1$
      & $\scnd{85.3}$ & \bluereb{$16.2$} & $63.6$ & $96.6$ \\
    \ourmodel-Large
      & $\scnd{85.2}$ & \bluereb{$\scnd{16.6}$} & $\best{59.3}$ & $\best{92.6}$
      & $\scnd{34.4}$ & \bluereb{$61.0$} & $\scnd{51.4}$ & $\best{76.3}$
      & $\best{88.2}$ & \bluereb{$\scnd{13.8}$} & $\best{73.3}$ & $\scnd{96.7}$
      & $\best{87.0}$ & \bluereb{$\scnd{15.0}$} & $\best{66.7}$ & $\scnd{97.2}$ \\
    \bottomrule
    \end{tabular}%
    }
\end{table*}

\subsection{Experimental Setup}
\label{ssec:experiments:setup}
\PAR{Data.} The training data is the combination of \bluereb{23} publicly available datasets: A2D2~\cite{geyer2020a2d2}, Argoverse2~\cite{2021argoverse2}, ARKit-Scenes~\cite{baruch2021arkitscenes}, BEDLAM~\cite{black2023bedlam}, BlendedMVS~\cite{yao2020blendedmvs}, DL3DV~\cite{ling2024dl3dv}, DrivingStereo~\cite{yang2019drivingstereo}, DynamicReplica~\cite{karaev2023dynamicreplica}, EDEN~\cite{le2021eden}, HOI4D~\cite{liu2022hoi4d}, HM3D~\cite{ramakrishnan2021habitat}, Matterport3D~\cite{chang2017matterport3d}, Mapillary-PSD~\cite{Lopez2020mapillary}, MatrixCity~\cite{li2023matrixcity}, MegaDepth~\cite{li2018megadepth}, NianticMapFree~\cite{arnold2022mapfree}, PointOdyssey~\cite{zheng2023pointodyssey}, ScanNet~\cite{dai2017scannet}, ScanNet++~\cite{yeshwanthliu2023scannetpp}, TartanAir~\cite{wang2020tartanair}, Taskonomy~\cite{zamir2018taskonomy}, Waymo~\cite{sun2020waymo}, and WildRGBD~\cite{xia2024wildrgbd} for a total of 16M images.
We evaluate the generalizability of models by testing them on \bluereb{ten} datasets not seen during training, grouped in different domains that are defined based on indoor, outdoor or \bluereb{``challenging'' settings}. 
The indoor group corresponds to the validation splits of SUN-RGBD~\cite{Song2015sunrgbd}, IBims~\cite{koch2022ibims}, and TUM-RGBD~\cite{sturm12tumrgbd}, the outdoor group comprises ETH3D~\cite{schoeps2017eth3d}, Sintel~\cite{Butler2012sintel}, DDAD~\cite{Guizilini2020ddad}, and NuScenes~\cite{nuscenes}, \bluereb{while the ``challenging'' domain is composed of HAMMER~\cite{jung2022hammer}, Booster~\cite{zamaramirez2024booster}, and FLSea~\cite{randall2023flsea}.}

\PAR{Evaluation Details.} All methods have been re-evaluated with a fair and consistent pipeline.
In particular, we do not exploit any test-time augmentations, and we utilize the same weights for all zero-shot evaluations.
We use the checkpoint corresponding to the zero-shot model for each method, \ie not fine-tuned on KITTI or NYU.
The metrics utilized in the main experiments are $\mathrm{\delta_1^{SSI}}$, $\mathrm{F_{A}}$, and $\mathrm{\rho_{A}}$.
$\mathrm{\delta_1}$ measures the depth estimation performance.
$\mathrm{F_{A}}$ is the area under the curve (AUC) of F1-score~\cite{ornek20222metrics} up to $1/20$ of the datasets' maximum depth and evaluates 3D estimation accuracy.
$\mathrm{\rho_{A}}$\bluereb{~\cite{Lee2021ctrlc,veicht2024geocalib}} evaluates the camera performance and is the AUC of the average angular error of camera rays up to 15$^{\circ}$.
We do not use parametric evaluation of \eg{}focal length, since it is a less flexible metric across diverse camera models and perfectly unrectified images.
Moreover, we present the fine-tuning ability of \ourmodel by training the final checkpoint on KITTI and NYU-Depth V2 and evaluating in-domain, as per standard practice.

\bluereb{Confidence predictions are evaluated with specific and established metrics such as \emph{AUSE}~\cite{poggi2020ause} (\wrt $\delta_1$) and its normalized version (\emph{nAUSE}) and \emph{Spearman’s $\rho$}.
$\mathrm{AUSE}$ ranks pixels by predicted uncertainty, progressively masks the most uncertain fraction, recomputes $\delta_1$ at each step, and integrates the gap to the oracle curve (ranking by true error).
Oracle's AUSE is $0$ by definition and $\mathrm{nAUSE}$ is $\mathrm{AUSE}$ normalized by the random error, \ie $\mathrm{nAUSE}=1$ equals \emph{random}.
$\rho$ is the rank correlation between predicted uncertainty and per-pixel error ($1$ perfect ordering, $0$ no monotonic relation, $<0$ inverted).
Both metrics therefore assess ranking quality: AUSE under progressive masking and $\rho$ as a global monotonicity check.}

\PAR{Implementation Details.} \ourmodel is implemented in PyTorch~\cite{pytorch} and CUDA~\cite{nickolls2008cuda}.
For training, we use the AdamW~\cite{Loshchilov2017adamw} optimizer ($\beta_1=0.9$, $\beta_2=0.999$) with an initial learning rate of $5\times{}10^{-5}$.
The learning rate is divided by a factor of 10 for the backbone weights for every experiment and weight decay is set to $0.1$.
We exploit Cosine Annealing as learning rate and weight decay scheduler to one-tenth starting from 30\% of the whole training.
\blue{We run 300k optimization iterations with a batch size of 128.
The training time amounts to 6 days on 16 NVIDIA 4090 with half precision.
The dataset sampling procedure follows a weighted sampler, where the weight of each dataset is its number of scenes.
Our augmentations are both geometric and photometric, \ie random resizing, cropping, and translation for the former type, and brightness, gamma, saturation, and hue shift for the latter.
We randomly sample the image ratio per batch between 2:1 and 1:2.}
Our ViT~\cite{Dosovitskiy2020VIT} backbone is initialized with weights from DINO-pretrained~\cite {oquab2023dinov2} models.
\bluereb{We train three models, one for each available ViT backbone, \ie Small, Base, and Large, corresponding to the UniDepthV2 variants ``-Small'', ``-Base'', and ``-Large''.}
For the ablations, we run 100k training steps with a ViT-S backbone, with the same training pipeline as for the main experiments.
\bluereb{Because the ablation runs use a shorter schedule and a fixed ViT-S backbone, the ablation ``final'' model is not directly comparable to the main UniDepthV2-Small results in \cref{tab:results:indoor,tab:results:outdoor,tab:results:challenge}.}

\subsection{Comparison with The State of The Art}
\label{ssec:experiments:comparison}

\begin{table*}[t]
\centering
\caption{\textbf{Results for Challenging Domains.} All methods are tested in a zero-shot fashion. Missing values (\textcolor{gray}{-}) indicate the model's inability to produce the respective output. \dag: Requires ground-truth (GT) camera for 3D reconstruction. \ddag: Requires GT camera for 2D depth map inference.}
\label{tab:results:challenge}
\vspace{-1em}
\resizebox{\linewidth}{!}{%
\begin{tabular}{l|cccc|cccc|cccc}
\toprule
\multirow{2}{*}{\textbf{Method}}  & \multicolumn{4}{c|}{HAMMER} & \multicolumn{4}{c|}{Booster} & \multicolumn{4}{c}{FLSea} \\
 & $\mathrm{\delta_1}\uparrow$ & $\mathrm{ARel}\downarrow$ & $\mathrm{F_A}\uparrow$ & $\mathrm{\rho_A}\uparrow$
 & $\mathrm{\delta_1}\uparrow$ & $\mathrm{ARel}\downarrow$ & $\mathrm{F_A}\uparrow$ & $\mathrm{\rho_A}\uparrow$
 & $\mathrm{\delta_1}\uparrow$ & $\mathrm{ARel}\downarrow$ & $\mathrm{F_A}\uparrow$ & $\mathrm{\rho_A}\uparrow$ \\
\midrule
Metric3D\textsuperscript{\dag \ddag}~\cite{yin2023metric3d}
 & $0.9$ & \bluereb{$122.5$} & \textcolor{gray}{-} & \textcolor{gray}{-}
 & \bluereb{1.2} & \bluereb{$201.2$} & \bluereb{\textcolor{gray}{-}} & \bluereb{\textcolor{gray}{-}}
 & \bluereb{0.7} & \bluereb{$117.9$} & \bluereb{\textcolor{gray}{-}} & \bluereb{\textcolor{gray}{-}} \\
Metric3Dv2\textsuperscript{\dag \ddag}~\cite{hu2024metric3dv2}
 & $\scnd{65.3}$ & \bluereb{$\scnd{21.4}$} & \textcolor{gray}{-} & \textcolor{gray}{-}
 & \bluereb{$9.1$} & \bluereb{$69.0$} & \bluereb{\textcolor{gray}{-}} & \bluereb{\textcolor{gray}{-}}
 & \bluereb{$17.5$} & \bluereb{$85.7$} & \bluereb{\textcolor{gray}{-}} & \bluereb{\textcolor{gray}{-}} \\
ZoeDepth\textsuperscript{\dag}~\cite{bhat2023zoedepth}
 & $0.9$ & \bluereb{$90.2$} & \textcolor{gray}{-} & \textcolor{gray}{-}
 & \bluereb{$28.5$} & \bluereb{$52.7$} & \bluereb{\textcolor{gray}{-}} & \bluereb{\textcolor{gray}{-}}
 & \bluereb{$15.7$} & \bluereb{$161.1$} & \bluereb{\textcolor{gray}{-}} & \bluereb{\textcolor{gray}{-}} \\
UniDepth~\cite{piccinelli2024unidepth}
 & $1.8$ & \bluereb{$56.9$} & $52.1$ & $55.3$
 & \bluereb{$42.9$} & \bluereb{$34.7$} & \bluereb{$42.7$} & \bluereb{$\scnd{73.7}$}
 & \bluereb{$29.9$} & \bluereb{$70.2$} & \bluereb{$53.6$} & \bluereb{$71.0$} \\
MASt3R~\cite{leroy2024master}
 & $2.2$ & \bluereb{$69.7$} & $38.1$ & $\best{86.5}$
 & \bluereb{$37.9$} & \bluereb{$46.1$} & \bluereb{$43.9$} & \bluereb{$69.6$}
 & \bluereb{$26.7$} & \bluereb{$41.5$} & \bluereb{$53.6$} & \bluereb{$67.5$} \\
DepthPro~\cite{bochkovskii2024depthpro}
 & $29.4$ & \bluereb{$37.5$} & $\scnd{71.0}$ & $69.1$
 & \bluereb{$58.4$} & \bluereb{$25.7$} & \bluereb{$11.1$} & \bluereb{\textcolor{gray}{-}}
 & \bluereb{$10.6$} & \bluereb{$145.2$} & \bluereb{$4.0$} & \bluereb{$4.7$} \\
\midrule
\ourmodel-Small
 & $30.6$ & \bluereb{$32.1$} & $57.0$ & $65.6$
 & \bluereb{$54.4$} & \bluereb{$30.3$} & \bluereb{$48.6$} & \bluereb{$70.1$}
 & \bluereb{$72.3$} & \bluereb{$20.6$} & \bluereb{$90.5$} & \bluereb{$87.2$} \\
\ourmodel-Base
 & $\best{89.7}$ & \bluereb{$22.4$} & $68.5$ & $\scnd{76.5}$
 & \bluereb{$\scnd{62.1}$} & \bluereb{$\scnd{24.4}$} & \bluereb{$\scnd{55.1}$} & \bluereb{$\best{75.2}$}
 & \bluereb{$\scnd{89.1}$} & \bluereb{$\scnd{12.2}$} & \bluereb{$\scnd{95.3}$} & \bluereb{$\best{93.1}$} \\
\ourmodel-Large
 & $\scnd{64.5}$ & \bluereb{$\best{17.5}$} & $\best{74.9}$ & $\scnd{78.3}$
 & \bluereb{$\best{67.6}$} & \bluereb{$\best{19.6}$} & \bluereb{$\best{57.9}$} & \bluereb{$66.6$}
 & \bluereb{$\best{90.5}$} & \bluereb{$\best{12.1}$} & \bluereb{$\best{95.9}$} & \bluereb{$\scnd{92.8}$} \\
\bottomrule
\end{tabular}%
}
\end{table*}

\begin{figure}[t]
    \centering
    \includegraphics[width=1.0\linewidth]{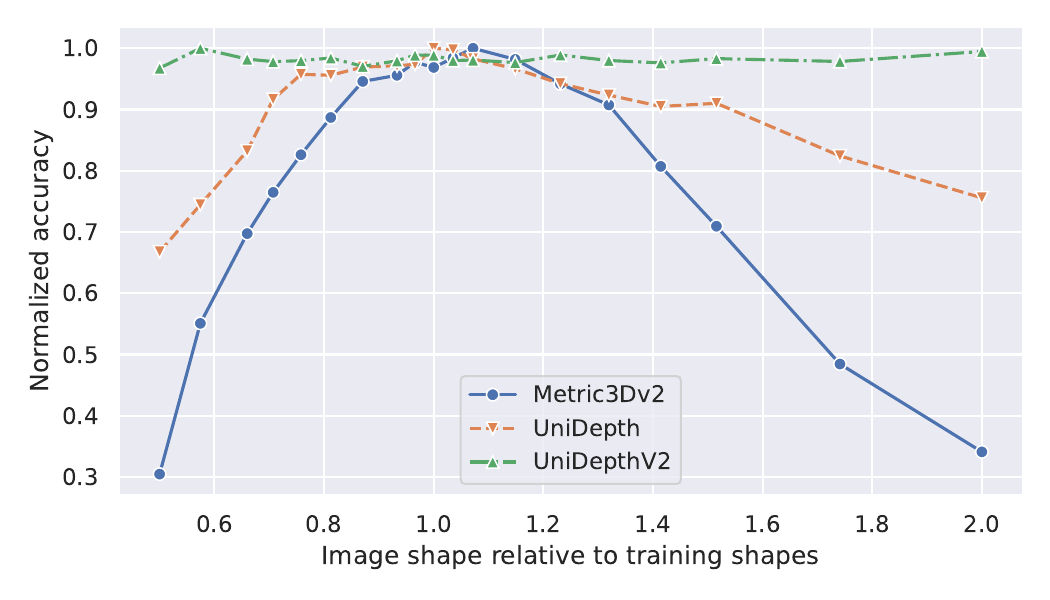}
    \vspace{-2em}
    \caption{\textbf{\bluereb{Insensitivity} to image shape.} \ourmodel is trained with a variable input shape pipeline in addition to random resizing for each of the image pairs. The proposed training strategy improves the robustness in terms of predicted depth scale and accuracy ($\delta_1$) to the input image's shape compared to two other state-of-the-art methods.}
    \label{fig:results:shape_invariance}
\end{figure}

We evaluate our method on \bluereb{a total of ten} zero-shot validation sets.
\bluereb{The domains cover indoor, outdoor, and challenging scenes, \eg underwater or transparent objects, shown in \Cref{tab:results:indoor,tab:results:outdoor,tab:results:challenge}, respectively.}
Our model performs better than or at least on par with all baselines, even outperforming methods that require ground-truth camera parameters at inference time, such as \cite{yin2023metric3d, hu2024metric3dv2}.
Notably, \ourmodel excels in 3D estimation, as reflected in the $\mathrm{F_A}$ metric, where it achieves a consistent improvement ranging from 0.5\% to 18.1\% over the second-best method. Additionally, it outperforms UniDepth~\cite{piccinelli2024unidepth} in nearly all cases, except for the $\mathrm{\rho_A}$ metric on IBims-1, DDAD, and NuScenes.
\bluereb{Moreover, we evaluate and compare inference efficiency in~\Cref{tab:results:efficiency}. All models are run with the same settings. UniDepthV2 is among the fastest and most efficient, even though it requires an additional module for camera prediction.}
This demonstrates that our proposed version is a significant step forward in both performance and efficiency.
However, the camera parameter estimation ($\mathrm{\rho_A}$) sees only marginal improvements, indicating that the limited diversity of training cameras remains a challenge that could be addressed with additional camera-only training, as suggested in~\cite{bochkovskii2024depthpro}.

\begin{table}[t]
\centering
\color{bluereb}
\caption{\textbf{Edge evaluation.} Scale-invariant boundary $\mathrm{F_1}$ on ETH3D, IBims-1, and Sintel following the protocol of \cite{bochkovskii2024depthpro}. \dag: inference performed at fixed resolution of $1536\times1536$ instead of the input original.}
\vspace{-1em}
\label{tab:results:edge_eval}
\begin{tabular}{l|c|c|c}
\toprule
\textbf{Method}  & ETH3D & IBims-1 & Sintel\\
\midrule
Metric3Dv2~\cite{hu2024metric3dv2} & $2.86$ & $13.48$ & $23.03$ \\
ZoeDepth~\cite{bhat2023zoedepth} & $0.75$ & $4.16$ & $2.63$ \\
UniDepth~\cite{piccinelli2024unidepth} & $0.22$ & $2.64$ & $0.42$ \\
MASt3R~\cite{leroy2024master} & $0.60$ & $1.45$ & $1.08$ \\
DepthPro\textsuperscript{\dag}~\cite{bochkovskii2024depthpro} & $\best{4.04}$ & $\best{19.38}$ & $\scnd{29.24}$ \\
\midrule
\ourmodel-Small & $2.13$ & $10.53$ & $20.58$ \\
\ourmodel-Base & $\scnd{2.99}$ & $12.93$ & $28.25$ \\
\ourmodel-Large & $2.95$ & $\scnd{13.75}$ & $\best{33.08}$ \\
\bottomrule
\end{tabular}
\end{table}

\Cref{tab:results:nyu_ft} and \Cref{tab:results:kitti_ft} show results for models fine-tuned on the NYU and KITTI training sets and evaluated on their respective validation splits, following standard protocols.
Fine-tuning performance serves as an indicator of a model's ability to specialize to specific downstream tasks and domains.
\ourmodel effectively adapts to new domains and outperforms methods \bluereb{with similar capacity} that were pre-trained on large, diverse datasets before fine-tuning on NYU or KITTI, such as~\cite{bhat2023zoedepth, hu2024metric3dv2, yang2024da2}\bluereb{.}
This is particularly evident in the outdoor setting (KITTI), as shown in \Cref{tab:results:kitti_ft}.
As detailed in \Cref{ssec:method:design}, our training strategy incorporates variable image aspect ratios and resolutions within the same distributed batch.
Combined with camera conditioning and invariance learning, this approach enhances the model’s robustness to changes in input image shape.
\Cref{fig:results:shape_invariance} quantifies this effect: the y-axis represents normalized metric accuracy ($\mathrm{\delta}_1$ scaled by the method’s maximum value), while the x-axis varies the image shape.
The normalization ensures a consistent scale across models.
\ourmodel is almost invariant to image shape, demonstrating that it can effectively trade off resolution for speed without sacrificing accuracy, as clearly illustrated in \Cref{fig:results:shape_invariance}.

\begin{figure}[t]
    \renewcommand{\arraystretch}{1}
    \centering
    \small
    \begin{tabular}{c}
        \includegraphics[width=0.99\linewidth]{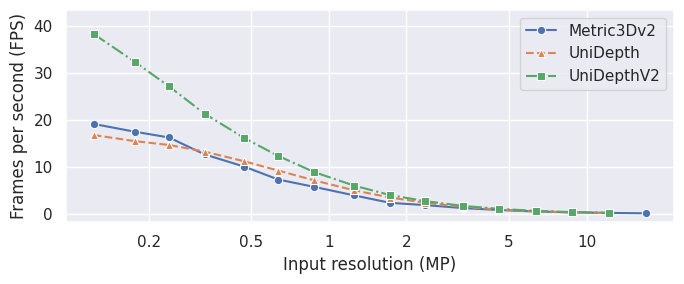}\\
        \includegraphics[width=0.99\linewidth]{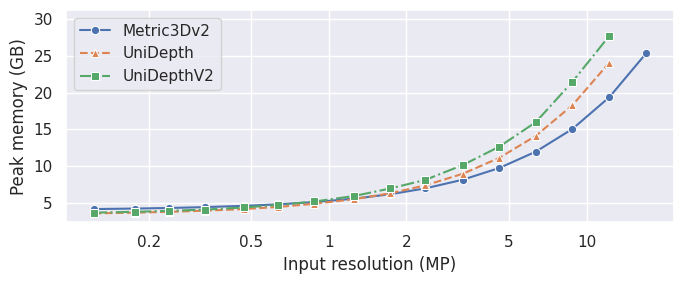}
    \end{tabular}
    \caption{\bluereb{\textbf{Impact of resolution on memory and runtime.} Frames-per-second (top) and peak memory (bottom) versus input resolution in Mega Pixels (log-scale). Missing points at 32MP are due to OOM.}}
    \label{fig:results:resol_impact}
\end{figure}

\bluereb{As shown in \Cref{fig:results:resol_impact}, increasing input resolution causes FPS to drop roughly inversely with resolution, while peak GPU memory grows near-quadratically. UniDepthV2 is around 2$\times$ faster than the baselines up to $\sim$2 Megapixel with a comparable memory footprint, but beyond $\sim$5 MP all methods become memory-bound ($>15$ GB) and converge to sub-FPS throughput. In practice, GPU memory sets the feasible operating point, motivating our shape-invariant inference that can trade resolution for speed without sacrificing accuracy (\Cref{fig:results:shape_invariance}).}
\bluereb{\Cref{tab:results:edge_eval} follows protocol from~\cite{bochkovskii2024depthpro}, but it is important to note a resolution asymmetry: DepthPro is evaluated at a fixed high input of $1536{\times}1536$, whereas other models that are more flexible \wrt input, such as \ourmodel, are run at the native input resolution, which is typically between one–sixth to one–quarter of DepthPro's.
Since boundary $\mathrm{F_1}$ improves with input resolution, \ie sharper and less aliased contours lead to higher true-positive matches, DepthPro’s scores are inflated by resolution. Even under this stricter setting, \ourmodel-Large attains the best result on Sintel and remains competitive on ETH3D and IBims-1 and clearly outperforms the original UniDepth.}

\begin{table}[t]
    \centering
    \caption{\textbf{Comparison on NYU validation set.} All models are trained on NYU. The first 4 are trained only on NYU. The last 4 are fine-tuned on NYU.}
    \vspace{-1em}
    \label{tab:results:nyu_ft}
    \resizebox{\columnwidth}{!}{
    \begin{tabular}{l|ccc|ccc}
        \toprule
        \multirow{2}{*}{\textbf{Method}} & $\mathrm{\delta}_{1}$ & $\mathrm{\delta}_{2}$ & $\mathrm{\delta}_{3}$ & $\mathrm{A.Rel}$ & $\mathrm{RMS}$ & $\mathrm{Log}_{10}$\\
         & \multicolumn{3}{c|}{\textit{Higher is better}} & \multicolumn{3}{c}{\textit{Lower is better}}\\
        \toprule
        BTS~\cite{Lee2019bts} & $88.5$ & $97.8$ & $99.4$ & $10.9$ & $0.391$ & $0.046$\\
        AdaBins~\cite{Bhat2020adabins} & $90.1$ & $98.3$ & $99.6$ & $10.3$ & $0.365$ & $0.044$\\
        NeWCRF~\cite{Yuan2022newcrf} & $92.1$ & $99.1$ & $\scnd{99.8}$ & $9.56$ & $0.333$ & $0.040$\\
        iDisc~\cite{piccinelli2023idisc} & $93.8$ & $99.2$ & $\scnd{99.8}$ & $8.61$ & $0.313$ & $0.037$\\
        ZoeDepth~\cite{bhat2023zoedepth} & $95.2$ & $\scnd{99.5}$ & $\scnd{99.8}$ & $7.70$ & $0.278$ & $0.033$\\
        Metric3Dv2~\cite{hu2024metric3dv2} & $\best{98.9}$ & $\best{99.8}$ & $\best{100}$ & $\scnd{4.70}$ & $\scnd{0.183}$ & $\best{0.020}$\\
        DepthAnythingv2~\cite{yang2024da2} & $98.4$ & $\mathbf{99.8}$ & $\mathbf{100}$ & $5.60$ & $0.206$ & $\scnd{0.024}$\\
        \midrule 
        \ourmodel\bluereb{-Large} & $\scnd{98.8}$ & $\best{99.8}$ & $\best{100}$ & $\best{4.68}$ & $\best{0.180}$ & $\best{0.020}$\\ 
        \bottomrule
    \end{tabular}}
\end{table}

\begin{table}[t]
    \centering
    \caption{\textbf{Comparison on KITTI Eigen-split validation set.} All models are trained on KITTI Eigen-split training and tested on the corresponding validation split. The first 4 are trained only on KITTI. The last 4 are fine-tuned on KITTI.}
    \vspace{-1em}
    \label{tab:results:kitti_ft}
    \resizebox{\columnwidth}{!}{
    \begin{tabular}{l|ccc|ccc}
        \toprule
        \multirow{2}{*}{\textbf{Method}} & $\mathrm{\delta}_{1}$ & $\mathrm{\delta}_{2}$ & $\mathrm{\delta}_{3}$ & $\mathrm{A.Rel}$ & $\mathrm{RMS}$ & $\mathrm{RMS}_{\log}$\\
         & \multicolumn{3}{c|}{\textit{Higher is better}} & \multicolumn{3}{c}{\textit{Lower is better}}\\
        \toprule
        BTS~\cite{Lee2019bts} & $96.2$ & $99.4$ & $99.8$ & $5.63$ & $2.43$ & $0.089$\\
        AdaBins~\cite{Bhat2020adabins} & $96.3$ & $99.5$ & $99.8$ & $5.85$ & $2.38$ & $0.089$\\
        NeWCRF~\cite{Yuan2022newcrf} & $97.5$ & $\scnd{99.7}$ & $\scnd{99.9}$ & $5.20$ & $2.07$ & $0.078$\\
        iDisc~\cite{piccinelli2023idisc} & $97.5$ & $\scnd{99.7}$ & $\scnd{99.9}$ &$5.09$ & $2.07$ & $0.077$\\
        ZoeDepth~\cite{bhat2023zoedepth} & $96.5$ & $99.1$ & $99.4$ & $5.76$ & $2.39$ & $0.089$ \\
        Metric3Dv2~\cite{yin2023metric3d} & $\scnd{98.5}$ & $\best{99.8}$ & $\best{100}$ & $\scnd{4.40}$ & $1.99$ & $\scnd{0.064}$\\
        DepthAnythingv2~\cite{yang2024da2} & $98.3$ & $\best{99.8}$ & $\best{100}$ & $4.50$ & $\scnd{1.86}$ & $0.067$ \\
        \midrule
        \ourmodel\bluereb{-Large} & $\best{98.9}$ & $\best{99.8}$ & $\scnd{99.9}$ & $\best{3.73}$ & $\best{1.71}$ & $\best{0.061}$ \\ 
        \bottomrule
    \end{tabular}}
\end{table}

\begin{figure*}[t]
    \renewcommand{\arraystretch}{1}
    \centering
    \small
    \hspace{-5pt}
    \begin{tabular}{cc|ccc}
        \multirow{1}{*}[0.5in]{\rotatebox[origin=c]{90}{KITTI}}
        & \includegraphics[width=0.22\linewidth]{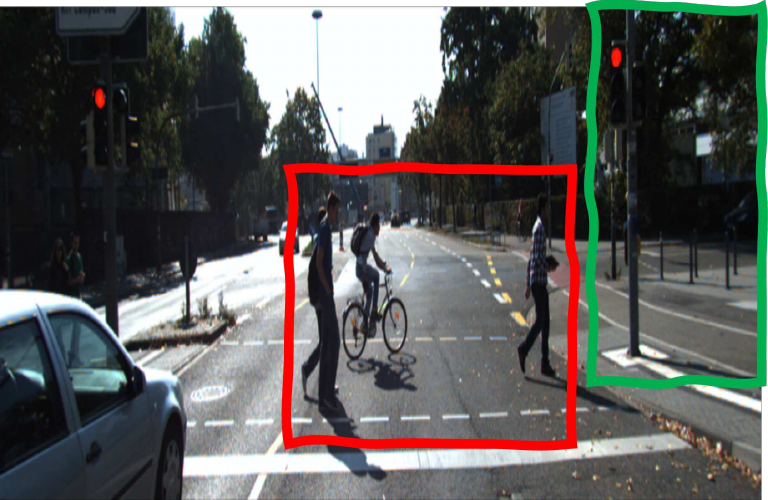}
        & \includegraphics[width=0.22\linewidth]{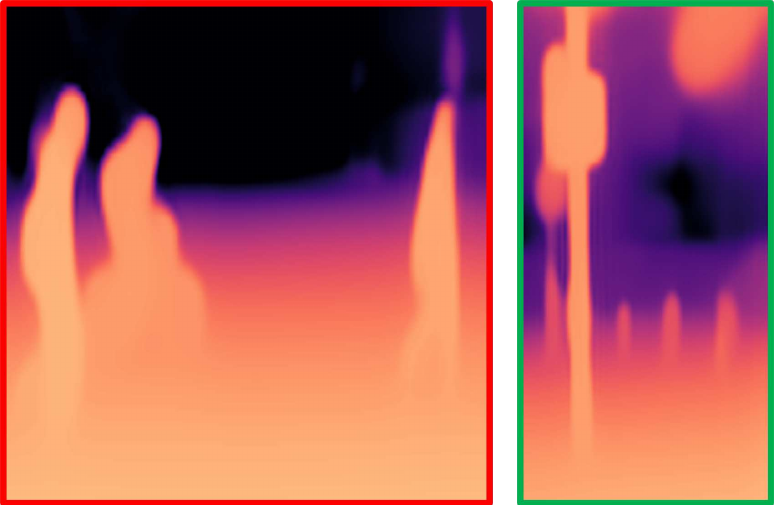}
        & \includegraphics[width=0.22\linewidth]{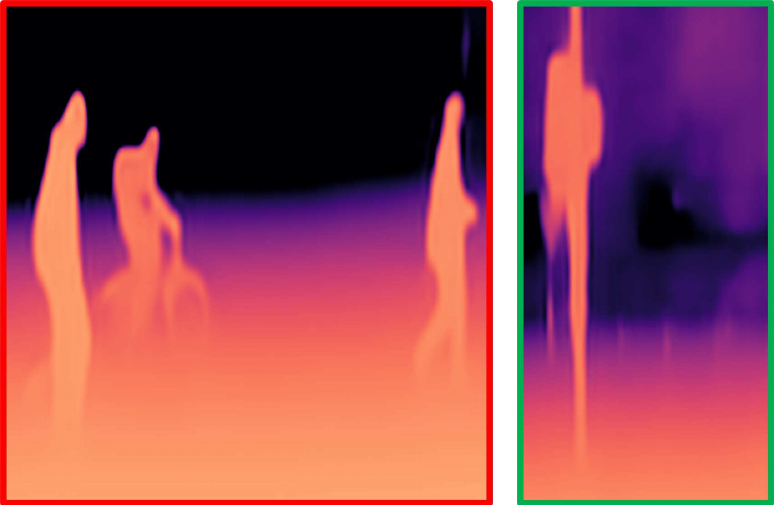}
        & \includegraphics[width=0.22\linewidth]{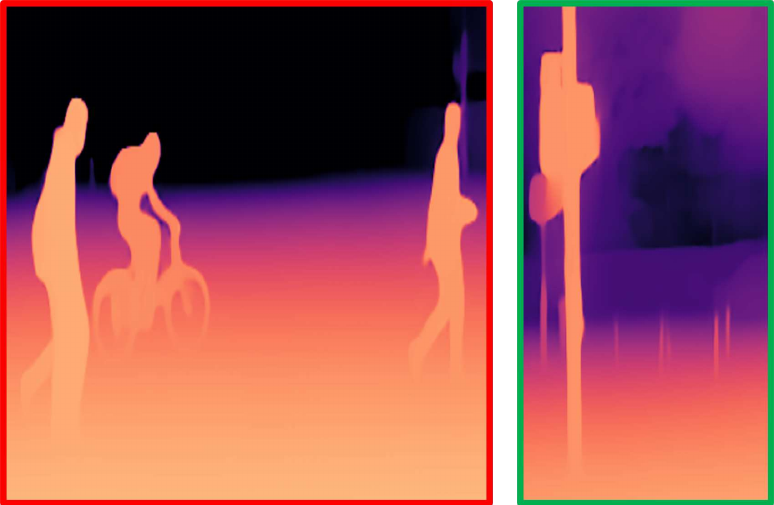}\\
        
        \multirow{1}{*}[0.5in]{\rotatebox[origin=c]{90}{Sintel}}
        & \includegraphics[width=0.22\linewidth]{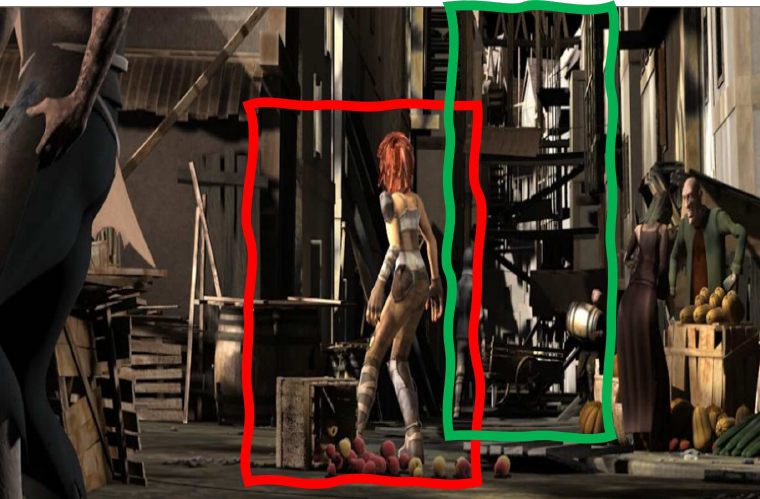}
        & \includegraphics[width=0.22\linewidth]{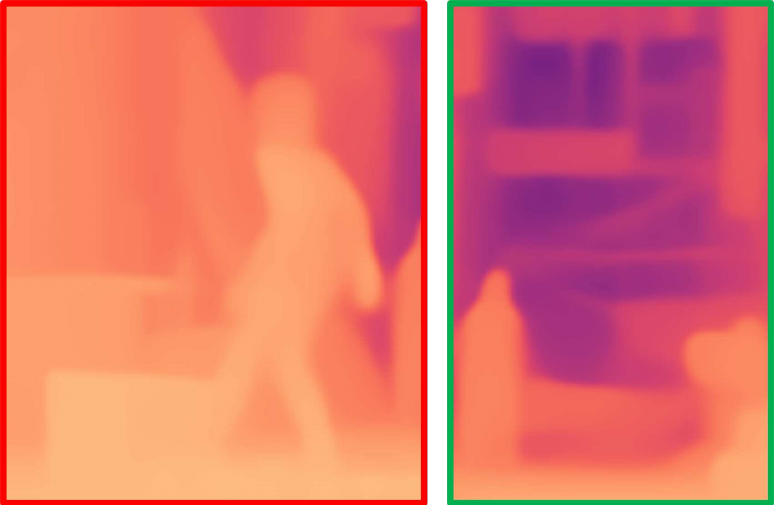}
        & \includegraphics[width=0.22\linewidth]{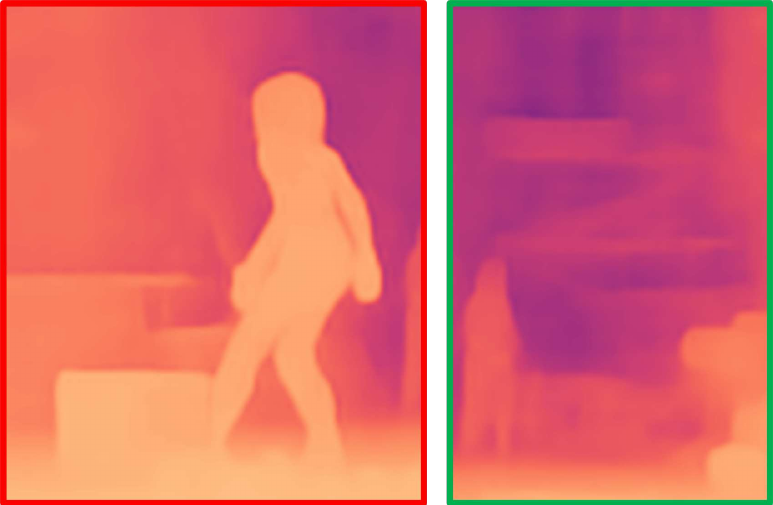}
        & \includegraphics[width=0.22\linewidth]{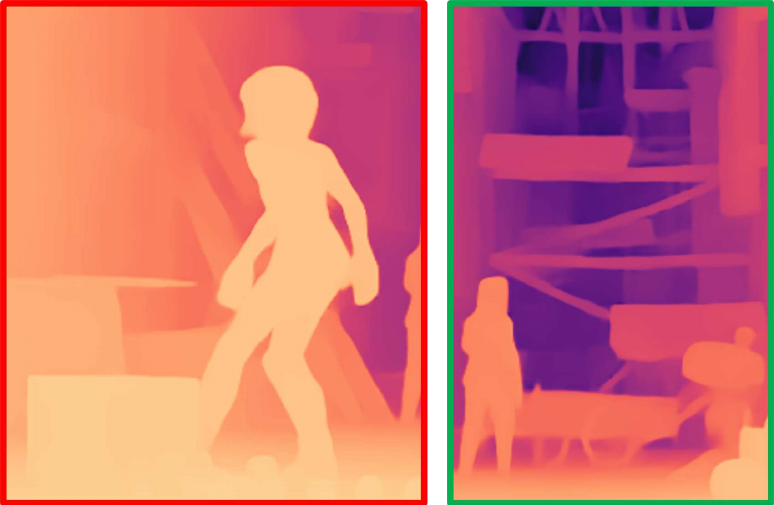}\\
        
        & RGB & UniDepth~\cite{piccinelli2024unidepth} & UniDepthV2 w/o $\mathcal{L}_\mathrm{EG-SSI}$ & \ourmodel\\
    \end{tabular}
    \vspace{-1em}
    \caption{\textbf{Comparisons of predicted edges.} Each row displays the input RGB image and the 2D depth maps predicted by compared methods, color-coded with the \textit{magma reverse} colormap with a range between 0 and 50 meters. Better viewed on a screen and zoomed in.
    }
    \label{fig:results:edges}
    \vspace{-1em}
\end{figure*}

\begin{table}[t]
\centering
\footnotesize
\color{bluereb}
\caption{\textbf{Uncertainty evaluation.} ``In-domain'' uses the validation splits of the training datasets; ``Zero-shot'' evaluates on unseen datasets. $\mathrm{AUSE}$~\cite{poggi2020ause} is the area under the sparsification error curve \wrt $\delta_1$ score, and $\mathrm{nAUSE}$ is the normalized version where 0 = oracle, 1 = random. $\rho$ is the Spearman rank correlation between predicted uncertainty and error.}
\vspace{-1em}
\label{tab:results:confidence}
\resizebox{\linewidth}{!}{\begin{tabular}{l|ccc|ccc}
    \toprule
    \multirow{2}{*}{\textbf{Model}} & \multicolumn{3}{c|}{{In-domain}} & \multicolumn{3}{c}{{Zero-shot}}\\
     & $\mathrm{AUSE} \downarrow$ & $\mathrm{nAUSE} \downarrow$ & $\rho \uparrow$ & $\mathrm{AUSE} \downarrow$ & $\mathrm{nAUSE} \downarrow$ & $\rho \uparrow$\\
    \midrule
    UniDepthV2-Small & $0.020$ & $0.221$ & $0.682$ & $0.040$ & $0.539$ & $0.289$ \\
    UniDepthV2-Base  & $0.018$ & $0.212$ & $0.721$ & $0.034$ & $0.637$ & $0.291$ \\
    UniDepthV2-Large & $0.017$ & $0.199$ & $0.744$ & $0.032$ & $0.645$ & $0.299$ \\
    \bottomrule
    \end{tabular}
}
\end{table}

\begin{table}[t]
    \centering
    \footnotesize
    \color{bluereb}
    \caption{\textbf{Efficiency results.} Inference efficiency results for all methods. Hardware is a A6000 with mixed precision and 0.5 Megapixel images, all methods use a ViT backbone; ViT-Large for competing methods. \dag: inference on native $1536 \times 1536$ resolution. \ddag: ConvNext-L backbone.}
    \label{tab:results:efficiency}
    \vspace{-1em}
    \resizebox{\linewidth}{!}{%
    \begin{tabular}{l|cccc}
    \toprule
    \textbf{Method} & Latency (ms) & Params (M) & FLOPS (T) & Memory (GiB)\\
    \midrule
    Metric3D\textsuperscript{\ddag}~\cite{yin2023metric3d} & $29.6$ & $203.2$ & $0.90$ & $1.71$ \\
    Metric3Dv2~\cite{hu2024metric3dv2}                     & $133.6$ & $411.9$ & $3.47$   & $3.50$ \\
    ZoeDepth~\cite{bhat2023zoedepth}                       & $64.8$  & $346.1$ & $2.08$   & $2.02$ \\
    UniDepth~\cite{piccinelli2024unidepth}                 & $91.0$  & $347.0$ & $2.02$   & $2.81$ \\
    MASt3R~\cite{leroy2024master}                          & $357.3$ & $688.6$ & $3.19$   & $4.94$ \\
    DepthPro\textsuperscript{\dag}~\cite{bochkovskii2024depthpro} & $270.6$ & $952.0$ & $19.3$  & $8.42$ \\
    \midrule
    \ourmodel-Small & $23.0$ & $34.18$  & $0.29$ & $0.66$ \\
    \ourmodel-Base  & $35.1$ & $114.4$ & $0.82$ & $1.32$ \\
    \ourmodel-Large & $65.4$ & $353.8$ & $2.17$ & $3.47$ \\
    \bottomrule
    \end{tabular}%
    }
\end{table}

\subsection{Ablation Studies}
\label{ssec:experiments:ablations}

The importance of each new component introduced in \ourmodel in \cref{sec:method} is evaluated by ablating the method in \blue{Tables \ref{tab:results:ablations_arch}, \ref{tab:results:ablations_loss}, \ref{tab:results:ablations_version}, and \ref{tab:results:ablations_repr}.}
All ablations exploit the predicted camera representation, if not stated otherwise.
\Cref{tab:results:ablations_arch} evaluates the impact of various architectural modifications compared to UniDepth~\cite{piccinelli2024unidepth}, analyzing their effects on both performance and efficiency.
\Cref{tab:results:ablations_loss} assesses the importance of the proposed loss function (\cref{ssec:method:egssi}) and examines the effect of applying the geometric invariance loss originally introduced in UniDepth~\cite{piccinelli2024unidepth} (\cref{ssec:method:consistency}) in different spaces.
The rationale behind our design choices is to maintain simplicity while maximizing effectiveness.
Additionally, in \Cref{tab:results:ablations_version} we analyze the role of camera conditioning and report results for the original UniDepth under the same training and evaluation setup as our method for a direct comparison.
The evaluation is based on four key metrics: $\mathrm{\delta}_1$, which measures metric depth accuracy; $\mathrm{SI}_{\log}$, which assesses scale-invariant scene geometry; $\mathrm{F_A}$, which captures the 3D estimation capability; and $\mathrm{\rho_A}$, which evaluates monocular camera parameter estimation.
All reported metrics correspond to the aggregated zero-shot performance across datasets, as detailed in \cref{ssec:experiments:setup}.

\begin{table}[t]
    \centering
    \caption{\textbf{Architectural ablations.} The different architectural additions (``+'') and subtractions (``-'') from the original UniDepth~\cite{piccinelli2024unidepth} are reported. ``- SHE + Sine'': camera encoding via Sine encoding instead of Spherical Harmonic Transform of the pinhole-based pencil of rays. ``- Attention'': attention layers in the decoder are removed. ``+ ResNet Blocks'': the attention layers in the decoder are substituted with simpler ResNet blocks. ``+ Multi-resol.'': the decoder has lateral connections with the shallower encoder layer, rather than a simpler merging of all resolutions in the bottleneck.}
    \vspace{-1em}
    \label{tab:results:ablations_arch}
    \resizebox{\linewidth}{!}{%
    \begin{tabular}{ll|cccc|cc}
    \toprule
    & \multirow{2}{*}{\textbf{Architecture}} & \multicolumn{4}{c|}{{Performance}} & \multicolumn{2}{c}{{Efficiency}} \\
     & & $\mathrm{\delta_1}\uparrow$ & $\mathrm{SI_{\log}}\downarrow$ & $\mathrm{F_A}\uparrow$ & $\mathrm{\rho_A}\uparrow$ & Latency \bluereb{(ms)}$\downarrow$ & Params \bluereb{(M)}$\downarrow$\\
    \midrule
    1 & UniDepth~\cite{piccinelli2024unidepth} & $54.5$ & $16.4$ & $56.1$ & $77.1$ & 73.2 & 35.2 \\
    2 & - SHE + Sine & $54.6$ & $16.4$ & $56.0$ & $76.9$ & 53.2 & 35.2 \\
    3 & - Attention & $50.3$ & $17.9$ & $51.0$ & $76.6$ & 20.4 & 29.0 \\
    4 & + ResNet Blocks & $52.6$ & $16.6$ & $55.0$ & $76.6$ & 24.0 & 33.5 \\
    5 & + Multi-resol & $54.5$ & $16.3$ & $56.0$ & $77.9$ & 25.0 & 34.2 \\
    \bluereb{6} & \bluereb{UniDepthV2} & $60.0$ & $15.3$ & $57.9$ & $79.8$ & 25.0 & 34.2 \\
    \bottomrule
    \end{tabular}%
    }
\end{table}
\begin{table}[t]
    \centering
    \caption{\textbf{Loss ablations.} $\mathcal{L}_{\mathrm{EG-SSI}}$ refers to either employing or not the proposed Edge-Guided Normalized loss; $\mathbf{O}_{\mathcal{L}_\mathrm{con}}$ indicates the output there the geometry consistency loss is applied to.}
    \vspace{-1em}
    \label{tab:results:ablations_loss}
    \resizebox{\linewidth}{!}{%
    \begin{tabular}{lcc|cccc}
    \toprule
    & \multirow{2}{*}{$\mathcal{L}_{\mathrm{EG-SSI}}$} & \multirow{2}{*}{$\mathbf{O}_{\mathcal{L}_\mathrm{con}}$} & \multicolumn{4}{c}{{Zero-shot Test}}\\
     & & & $\mathrm{\delta_1}\uparrow$ & $\mathrm{SI_{\log}}\downarrow$ & $\mathrm{F_A}\uparrow$ & $\mathrm{\rho_A}\uparrow$\\
    \midrule
    1 & \xmark & $\mathbf{D|E}$ & $54.5$ & $16.3$ & $56.0$ & $77.9$\\
    2 & \xmark & $\mathbf{Z}$   & $55.3$ & $16.2$ & $56.1$ & $78.2$\\
    3 & \cmark & $\mathbf{Z}$   & $60.0$ & $15.3$ & $57.9$ & $79.8$\\
    \bottomrule
    \end{tabular}%
    }
\end{table}

\begin{table}[t]
    \centering
    \caption{\textbf{Model ablations.} The ``Model'' column refers to architecture and training strategy employed. ``V1'' is the original UniDepth, while ``V2'' is the proposed \ourmodel. ``Cond'' specifies whether the camera-prompting mechanism is present or not.}
    \vspace{-1em}
    \label{tab:results:ablations_version}
    \resizebox{\linewidth}{!}{%
    \begin{tabular}{lcc|cccc}
    \toprule
    & \multirow{2}{*}{\textbf{Model}} & \multirow{2}{*}{\textbf{Cond}} & \multicolumn{4}{c}{{Zero-shot Test}}\\
     & & & $\mathrm{\delta_1}\uparrow$ & $\mathrm{SI_{\log}}\downarrow$ & $\mathrm{F_A}\uparrow$ & $\mathrm{\rho_A}\uparrow$\\
    \midrule
    1 & V1 & \xmark & $50.1$ & $18.0$ & $50.8$ & $76.7$\\
    2 & V1 & \cmark & $54.5$ & $16.4$ & $56.1$ & $77.1$\\
    3 & V2 & \xmark & $49.3$ & $18.4$ & $49.2$ & $76.6$\\
    4 & V2 & \cmark & $54.5$ & $16.3$ & $56.0$ & $77.9$\\
    \bottomrule
    \end{tabular}%
    }
\end{table}

\begin{table}[t]
    \centering
    \color{bluereb}            
    \caption{\textbf{Representation.} The ``Repr'' column refers to direct Cartesian regression (\textit{xyz}) regression \vs UniDepth's pseudo-spherical output (\textit{sph}). The ``Model'' column refers to architecture and training strategy employed: ``V1'' is the original UniDepth, while ``V2'' is the proposed \ourmodel.}
    \vspace{-1em}
    \label{tab:results:ablations_repr}
    \resizebox{\linewidth}{!}{%
    \begin{tabular}{lcc|cccc}
    \toprule
    & \multirow{2}{*}{\textbf{Model}} & \multirow{2}{*}{\textbf{Repr}} & \multicolumn{4}{c}{{Zero-shot Test}}\\
     & & & $\mathrm{\delta_1}\uparrow$ & $\mathrm{SI_{\log}}\downarrow$ & $\mathrm{F_A}\uparrow$ & $\mathrm{\rho_A}\uparrow$\\
    \midrule
    1 & V1 & xyz & $48.2$ & $21.2$ & $41.2$ & $61.3$\\
    2 & V1 & sph & $54.5$ & $16.4$ & $56.1$ & $77.1$\\
    3 & V2 & xyz & $49.1$ & $20.4$ & $38.8$ & $60.2$\\
    4 & V2 & sph & $60.0$ & $15.3$ & $57.9$ & $79.8$\\
    \bottomrule
    \end{tabular}%
    }
\end{table}

\PAR{Architecture.} \Cref{tab:results:ablations_arch} outlines the key modifications that transform the original UniDepth~\cite{piccinelli2024unidepth} architecture into \ourmodel.
The first major change is the removal of spherical harmonics (SH)-based encoding, which is computationally inefficient.
Instead, we revert to standard Sine encoding (row 2).
While the difference in performance is minimal in our setup, we hypothesize that the encoding’s impact diminishes as the model benefits from larger and more diverse training data across different cameras.
Next, we eliminate the attention mechanism in row 3 due to its high computational cost.
This removal results in a significant performance drop, \eg -4.3\% for $\mathrm{\delta}_1$, but yields a greater than 2x improvement in efficiency.
In row 4, we replace the pure MLP-based decoder with ResNet blocks, introducing spatial $3\times3$ convolutions.
This modification enhances performance by leveraging local spatial structure while inducing a minimal impact on efficiency.
Finally, row 5 integrates a multi-resolution feature fusion from the encoder to the decoder, following an FPN-style design.
This final architecture significantly reduces computational cost while preserving overall performance: the final model (row~5) achieves similar performance to the original UniDepth (row~1) while requiring only one-third of the computation. \bluereb{Row~6 reports the full \ourmodel configuration, \ie the architecture from row~5 with training augmented with the proposed losses.}
\PAR{$\mathcal{L}_{\mathrm{EG-SSI}}$ Loss.} The effectiveness of the proposed $\mathcal{L}_{\mathrm{EG-SSI}}$ loss, detailed in \cref{ssec:method:egssi}, is evaluated in row 2 \vs row 3 of \Cref{tab:results:ablations_loss}.
Introducing this loss results in a 4.7\% improvement in $\mathrm{\delta}_1$ and a 1.8\% improvement in $\mathrm{F_A}$, demonstrating its contribution to both metric accuracy and 3D estimation.
Interestingly, despite $\mathcal{L}_{\mathrm{EG-SSI}}$ not explicitly supervising camera parameter estimation, the $\mathrm{\rho_A}$ metric also shows improvement.
This suggests that the loss contributes to a less noisy training process, leading to better feature representations in the encoder.
A qualitative comparison of the impact of $\mathcal{L}_{\mathrm{EG-SSI}}$ is presented in \cref{fig:results:edges}.
The difference between the third and fourth columns highlights the visual impact of the proposed loss, particularly in refining depth discontinuities.
Additionally, the comparison between the second and third columns illustrates the combined effect of architectural changes and increased data diversity, showing improved reconstruction of finer details, such as body parts that were previously smoothed or missed.
\PAR{$\mathcal{L}_{\mathrm{con}}$ Output Space.} \ourmodel introduces multiple instances of camera-conditioned depth features $\mathbf{D}|\mathbf{E}$, corresponding to different decoder resolutions, as described in \cref{ssec:method:design}.
This contrasts with the original UniDepth~\cite{piccinelli2024unidepth}, which relied on a single conditioning point.
Given this architectural shift, we argue that deep conditioning may not be optimal.
Features at different resolutions encode varying levels of abstraction, and enforcing deep conditioning introduces additional design freedom.
\Cref{tab:results:ablations_loss} investigates where to apply the consistency loss ($\mathcal{L}_{\mathrm{con}}$) from~\cite{piccinelli2024unidepth}: either directly in the output space ($\mathbf{Z}$, row 2) or within the camera-conditioned features at each scale ($\mathbf{D}|\mathbf{E}$, row 1).
The results indicate minimal differences from applying the loss directly in the output space. Therefore, based on Occam's razor, we adopt the simpler and more effective design from row 2 as the final approach.
\PAR{Conditioning Impact.} As previously explored in~\cite{piccinelli2024unidepth}, we analyze the impact of our proposed camera conditioning in \Cref{tab:results:ablations_version}.
This ablation includes both UniDepth and \ourmodel under the same conditions—without $\mathcal{L}_{\mathrm{EG-SSI}}$ and without invariance applied to deep features ($\mathbf{D}|\mathbf{E}$).
The results show that conditioning has an even stronger positive effect for \ourmodel, as evidenced by comparing row 3 \vs row 4 against the comparison of row 1 \vs row 2.
\bluereb{\PAR{Camera Disentanglement.} We re-evaluate the output representation for \ourmodel in \cref{tab:results:ablations_repr}, mirroring the analysis in~\cite{piccinelli2024unidepth}. With otherwise matched camera settings, we observe a similar comparative picture in the \ourmodel setting: a significant benefit of our pseudo-spherical representation for depth-specific metrics and a substantial benefit for 3D and camera accuracy.}
\bluereb{\PAR{Confidence.} We evaluate in \cref{tab:results:confidence} the confidence estimator introduced in \cref{ssec:method:design} on the validation splits of the training datasets (``In-domain'') and on unseen datasets (``Zero-shot'').
In-domain, uncertainty aligns well with error: $\mathrm{nAUSE}$ is low ($0.199$–$0.221$) and $\rho$ is high ($0.68$–$0.74$), both improving as model capacity is increased.
Under domain shift, the quality drops, \ie $\mathrm{nAUSE}$ rises to $0.54$–$0.65$ and $\rho$ falls to $0.29$, \ie indicating that ranking is partly preserved but calibration degrades.
The opposing trend, namely slightly higher $\rho$ yet worse $\mathrm{nAUSE}$ for larger models, is probably capacity-driven overfitting: bigger models learn sharper, edge-focused uncertainty priors that order local errors correctly while misestimating their magnitude in novel domains.
Nevertheless, even zero-shot uncertainty remains informative and far from random, enabling reliability-aware masking and exploiting our resolution/speed trade-offs at inference.}

\newlength{\pairsep}\setlength{\pairsep}{6mm}        
\newlength{\pairrowsep}\setlength{\pairrowsep}{6mm}  

\newlength{\pairW}\setlength{\pairW}{\dimexpr(\textwidth-2\pairsep)/3\relax} 
\newlength{\imgW}\setlength{\imgW}{0.5\pairW}   
\newlength{\imgH}\setlength{\imgH}{25mm}

\begin{figure*}[t]
  \centering
  \setlength{\tabcolsep}{0pt} 
  \begin{tabular}{@{}c@{\hspace{\pairsep}}c@{\hspace{\pairsep}}c@{}}

    \makebox[\pairW][l]{%
      \includegraphics[width=\imgW,height=\imgH,keepaspectratio=false]{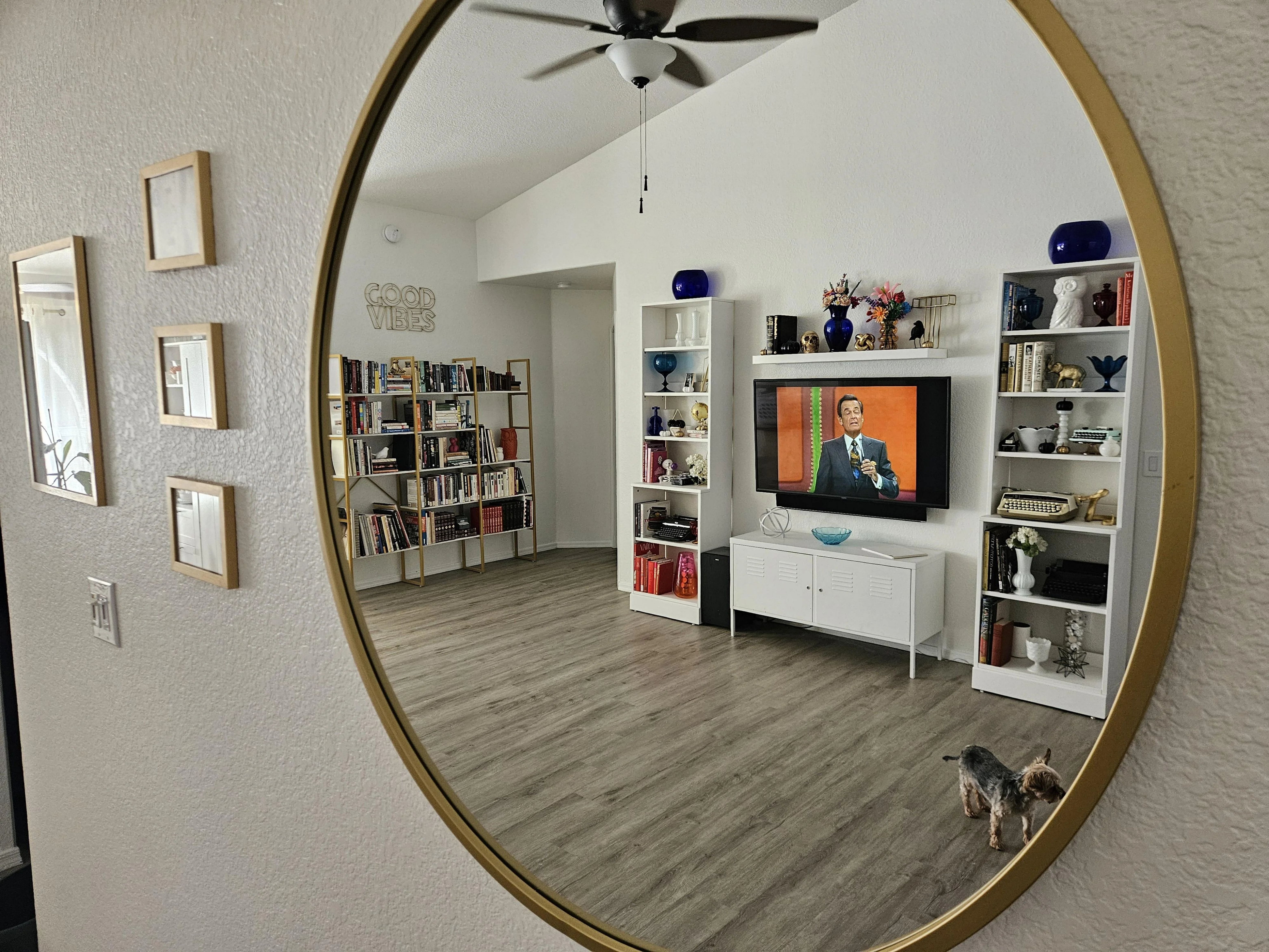}%
      \includegraphics[width=\imgW,height=\imgH,keepaspectratio=false]{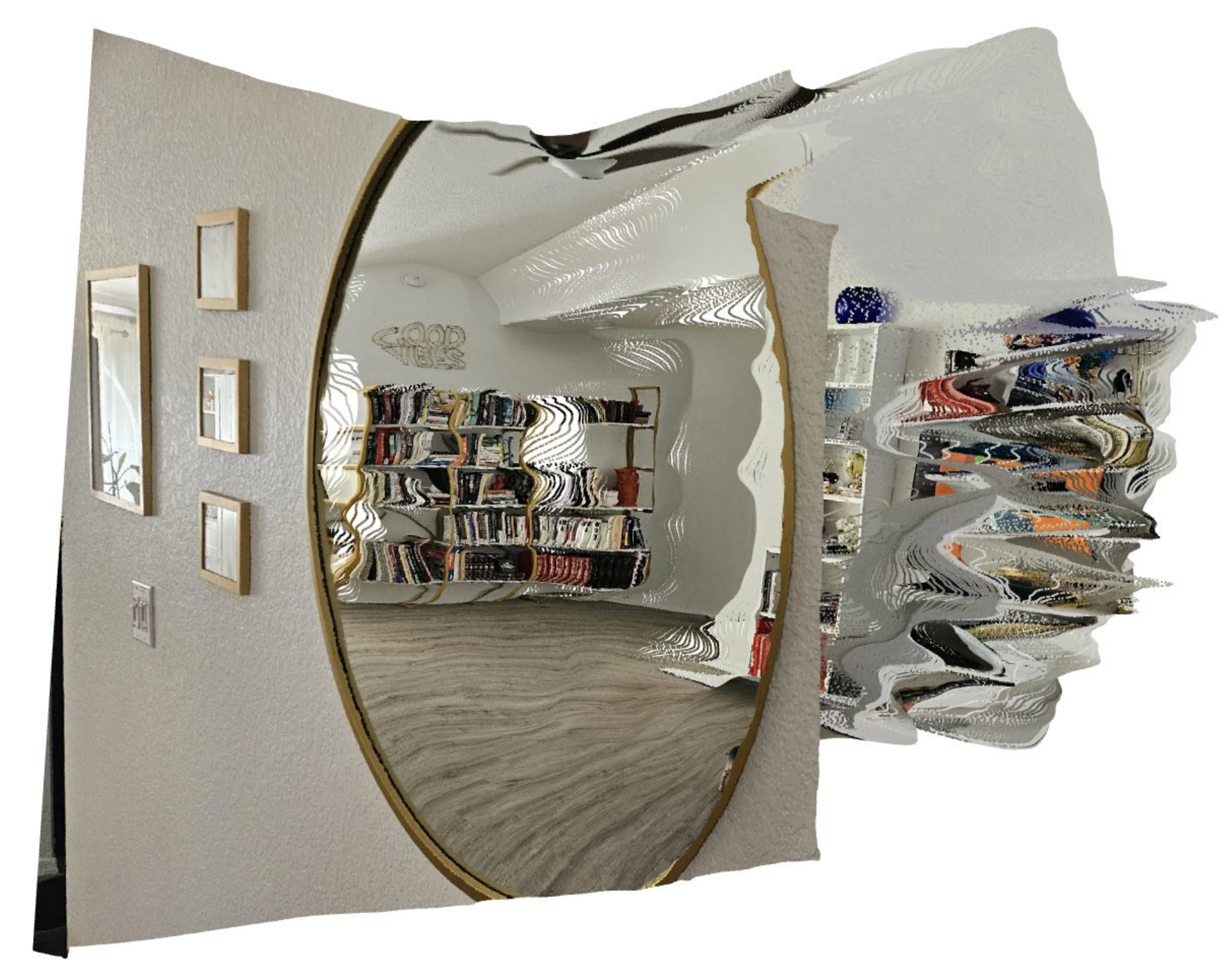}%
    } &
    \makebox[\pairW][l]{%
      \includegraphics[width=\imgW,height=\imgH,keepaspectratio=false]{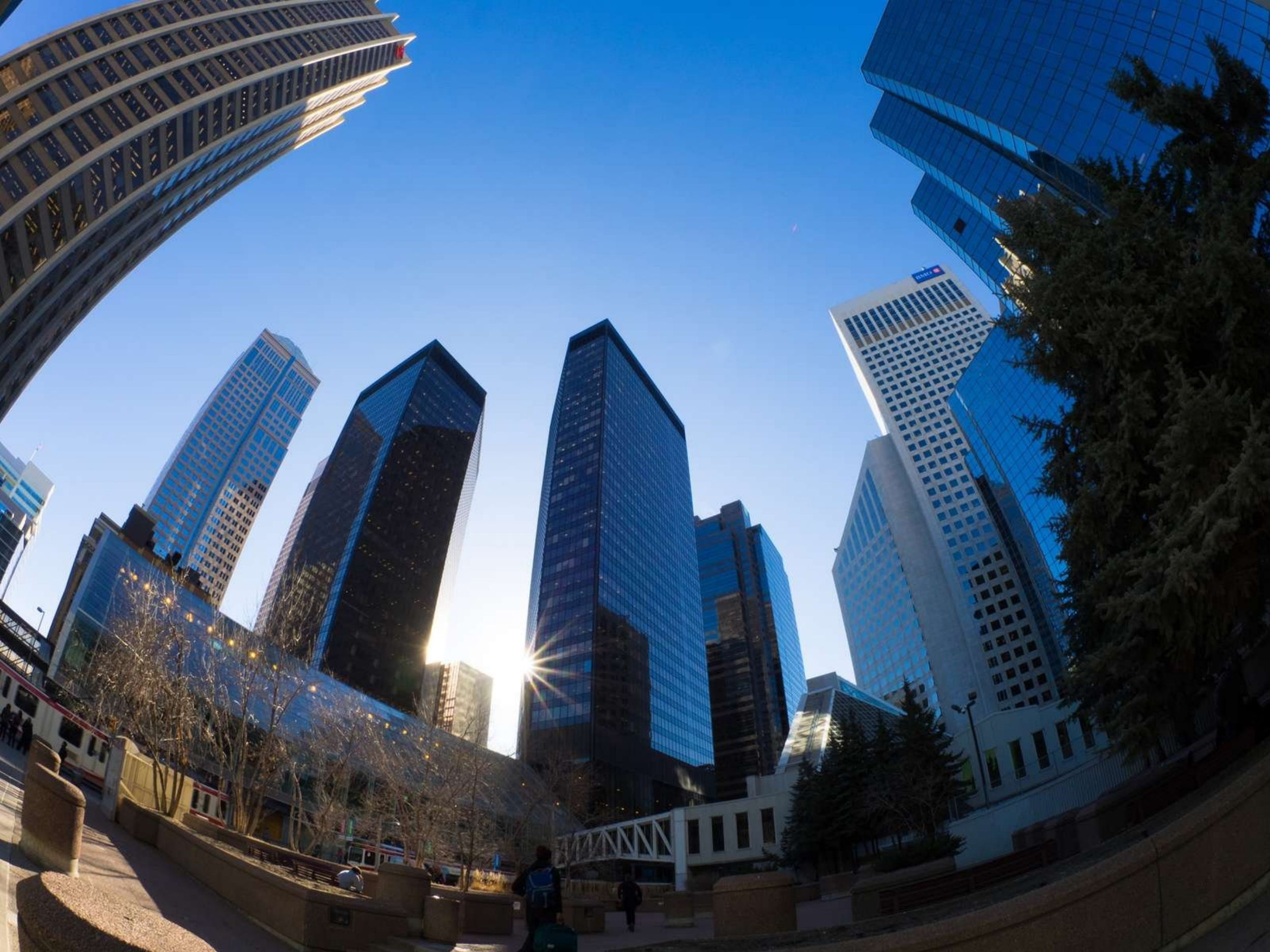}%
      \includegraphics[width=\imgW,height=\imgH,keepaspectratio=false]{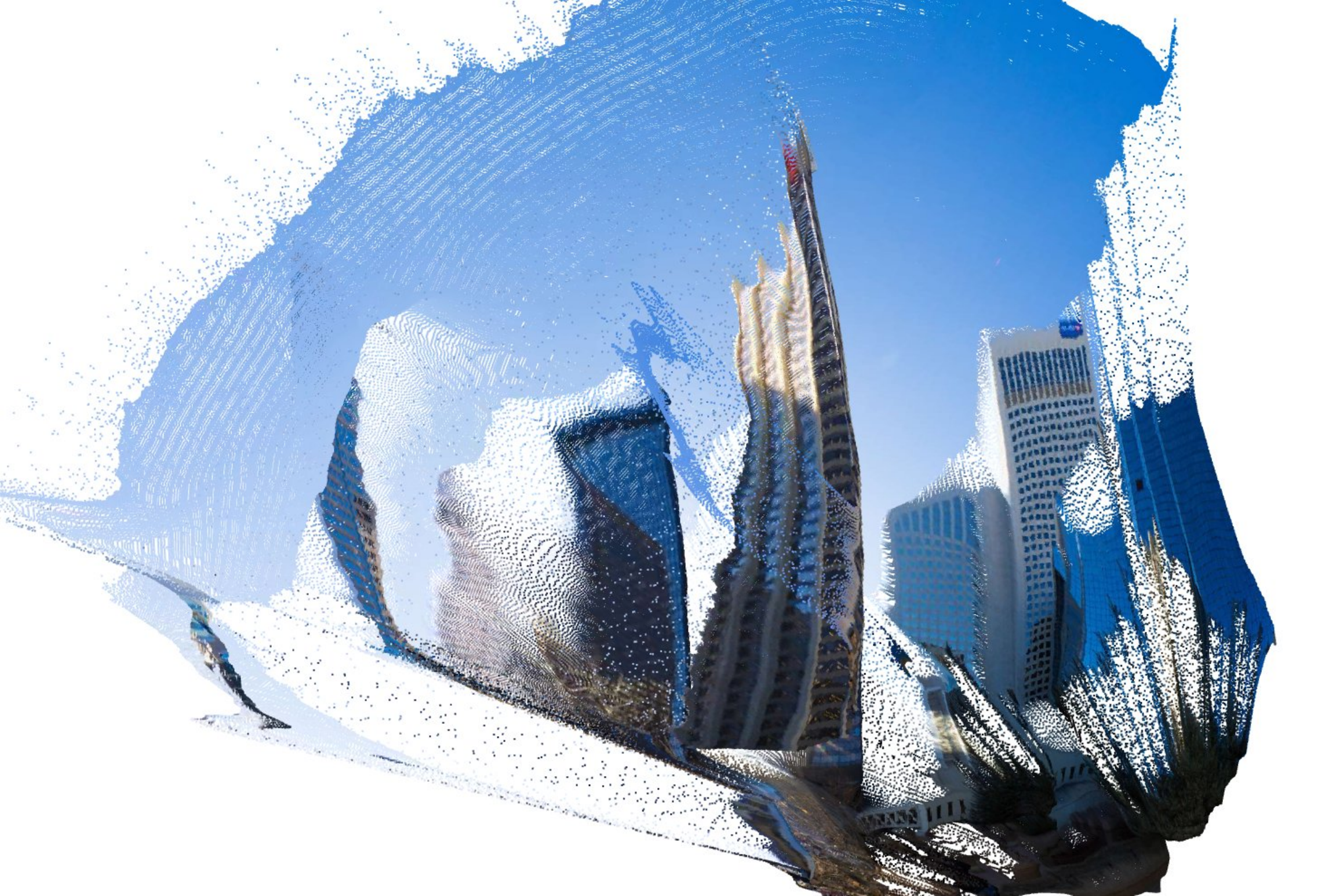}%
    } &
    \makebox[\pairW][l]{%
      \includegraphics[width=\imgW,height=\imgH,keepaspectratio=false]{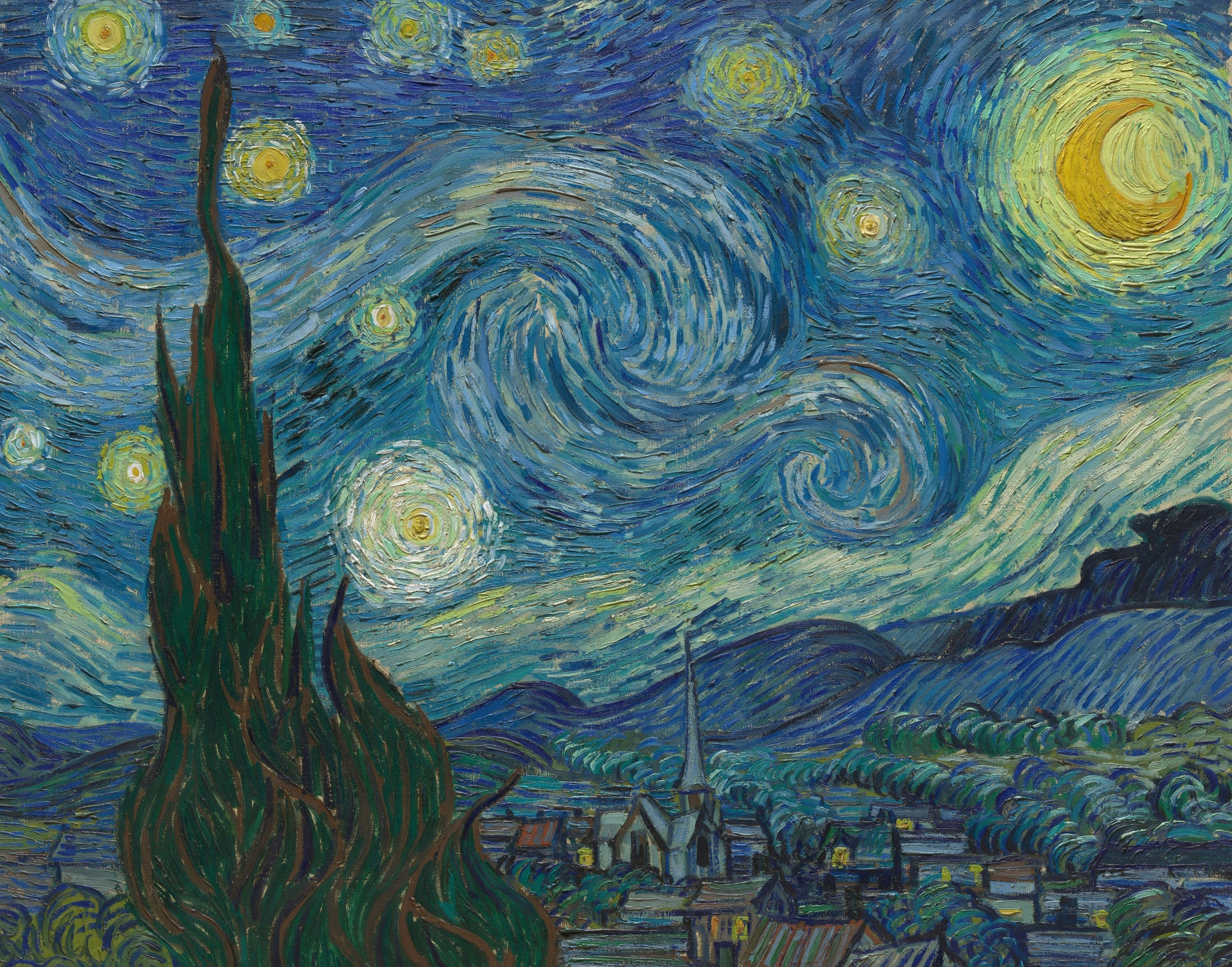}%
      \includegraphics[width=\imgW,height=\imgH,keepaspectratio=false]{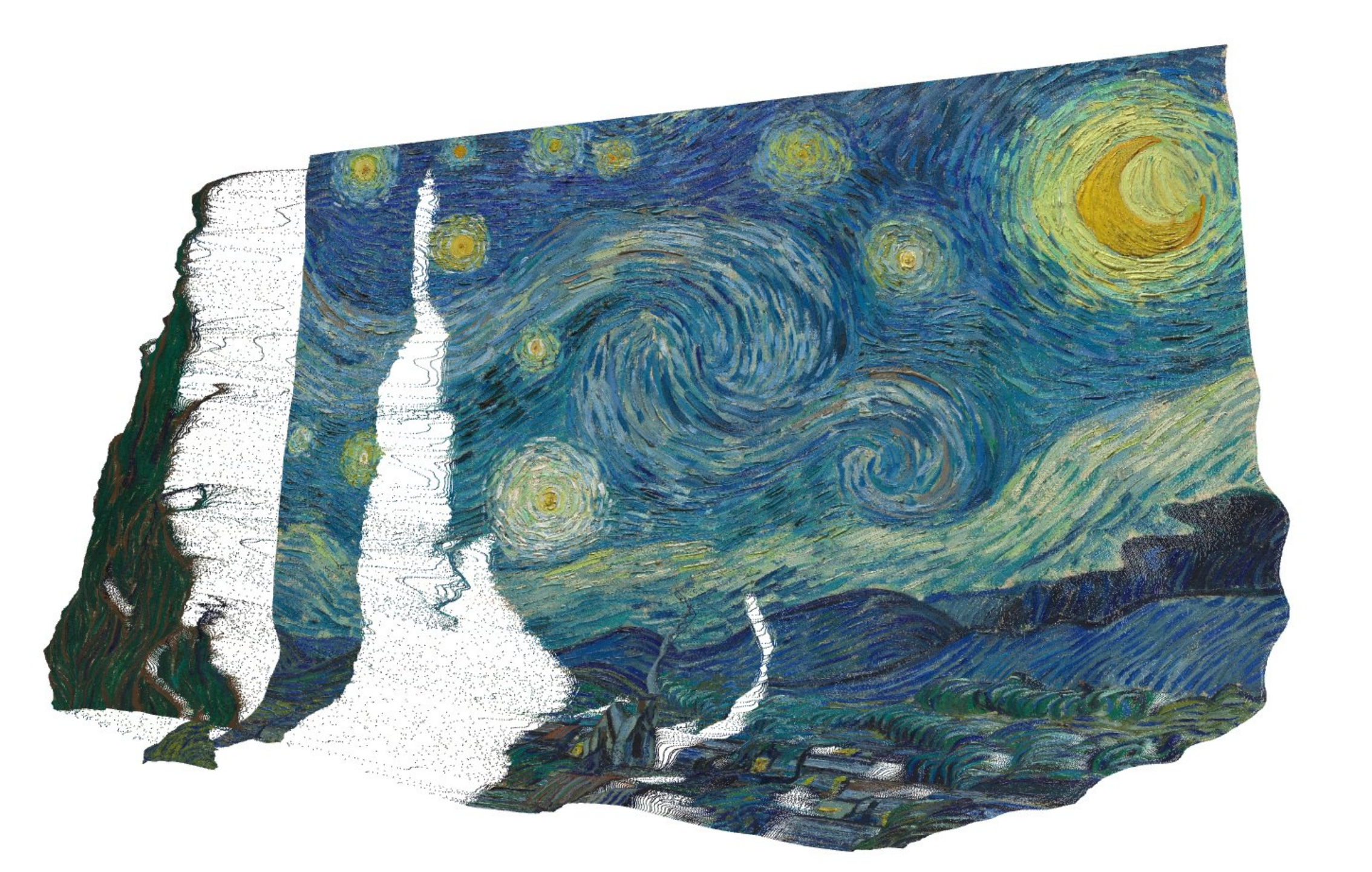}%
    } \\[\pairrowsep]

    \makebox[\pairW][l]{%
      \includegraphics[width=\imgW,height=\imgH,keepaspectratio=false]{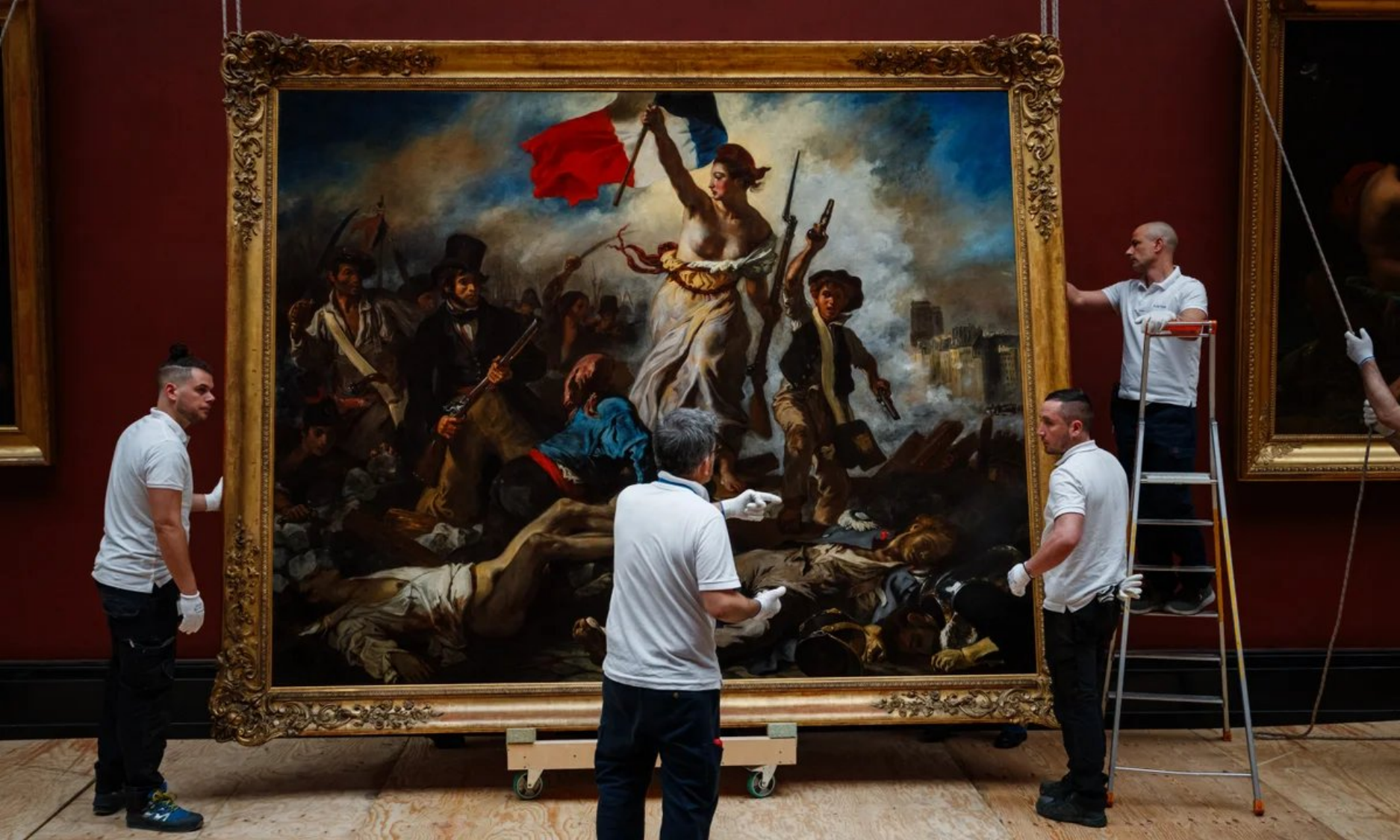}%
      \includegraphics[width=\imgW,height=\imgH,keepaspectratio=false]{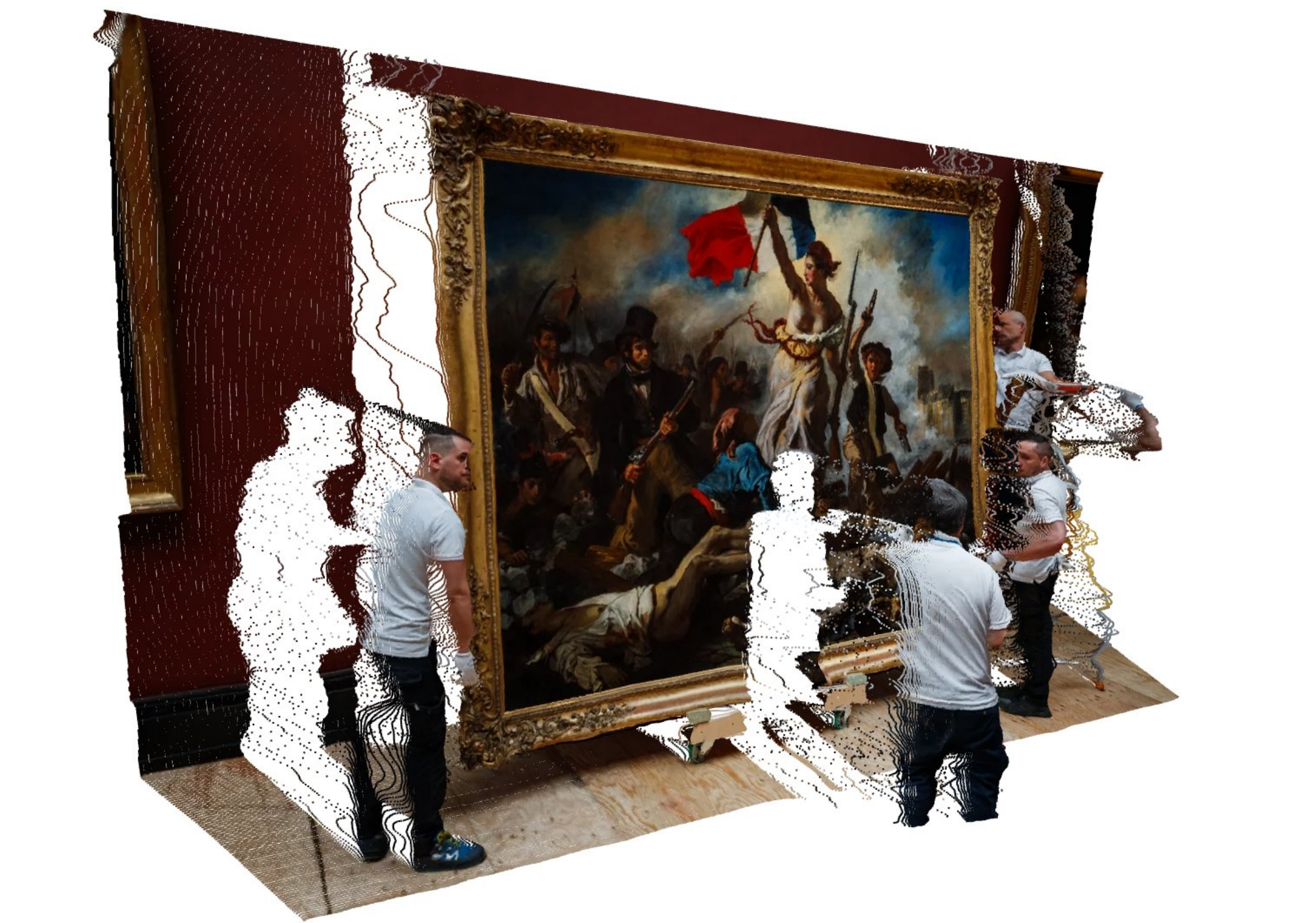}%
    } &
    \makebox[\pairW][l]{%
      \includegraphics[width=\imgW,height=\imgH,keepaspectratio=false]{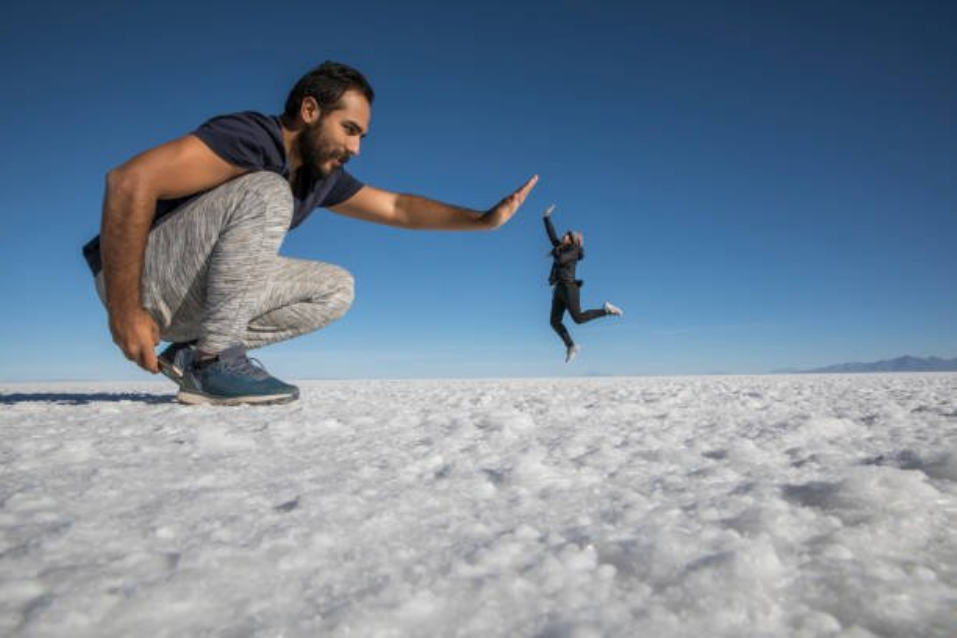}%
      \includegraphics[width=\imgW,height=\imgH,keepaspectratio=false]{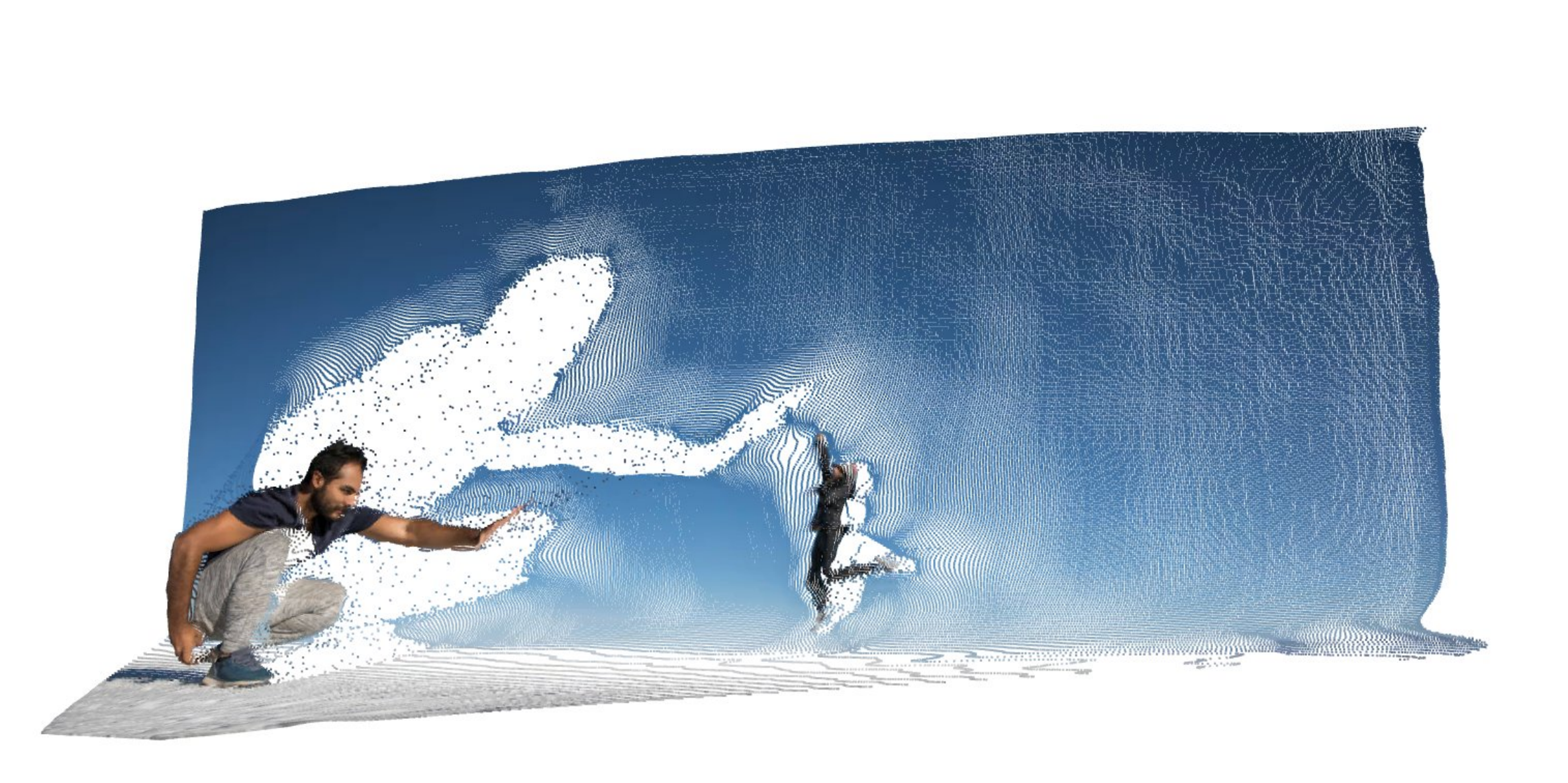}%
    } &
    \makebox[\pairW][l]{%
      \includegraphics[width=\imgW,height=\imgH,keepaspectratio=false]{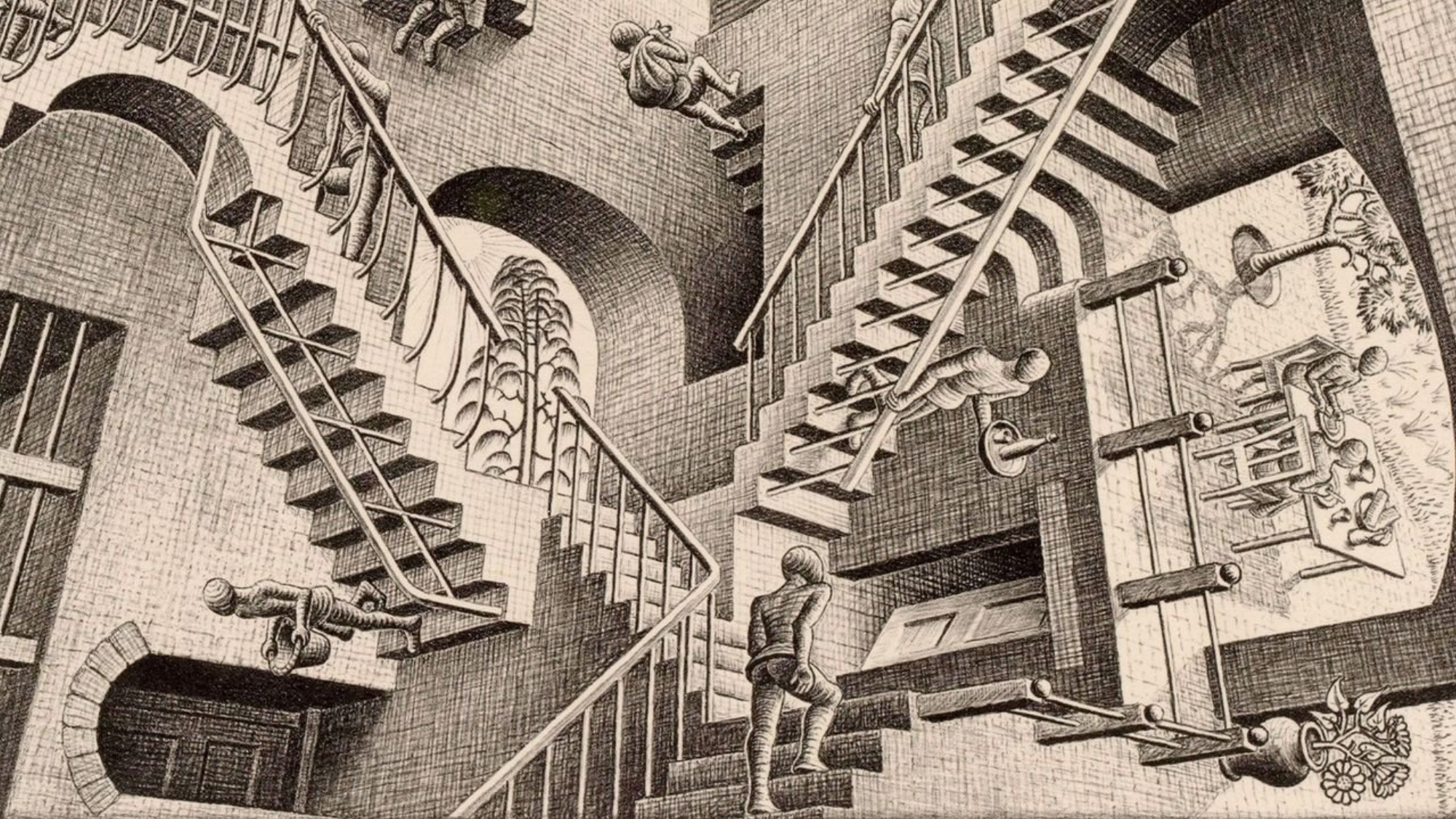}%
      \includegraphics[width=\imgW,height=\imgH,keepaspectratio=false]{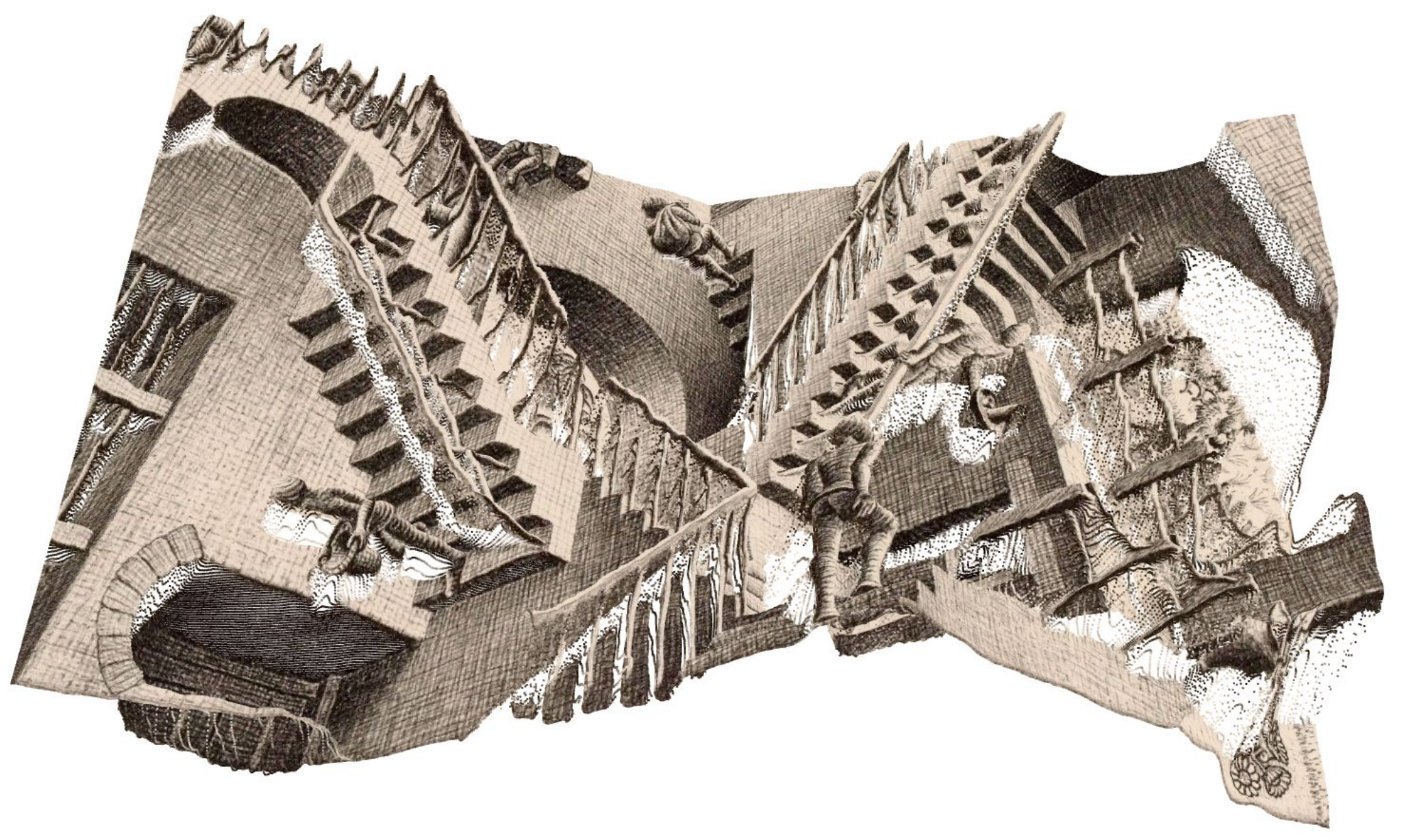}%
    } \\

  \end{tabular}

  \caption{\bluereb{\textbf{Failure and Edge Cases.} \ourmodel is run on challenging images collected from the internet. The domains include mirrors, non-pinhole cameras, paintings, and optical illusions, both from cameras and drawings.}}
  \label{fig:results:failures}
  \vspace{-1em}
\end{figure*}

\bluereb{\subsection{Failure Cases and Limitations}}

\bluereb{We present in~\cref{fig:results:failures} six images from the internet that pose unusual, challenging conditions for depth models, including mirrors, paintings, optical illusions, and non-pinhole projection.
While \ourmodel does not exhibit catastrophic failures on these examples, it can resolve certain human-related optical illusions and correctly recognize a painting when its frame and contextual objects are visible (\eg ``Liberty Leading the People'').
However, it struggles with non-realistic paintings and distorted geometry (\eg ``Starry Night''), though it still separates foreground from background.
Mirrors remain particularly challenging and ambiguous: without sufficient context, the model often interprets them as just a cavity and accommodates strong deformation.
In addition, \ourmodel struggles to represent non-pinhole projection and is not able to rectify the deformation as the implicit camera representation is pinhole-based as described in \cref{ssec:method:camera_module}.}
\vspace{1em}

\section{Conclusion}
\label{sec:conclusion}

\blue{We introduced \ourmodel, a universal monocular metric depth estimation model that enhances generalization across diverse domains without requiring camera parameters at test time.
By improving both the model architecture and introducing new loss functions in the training objective, \ourmodel achieves state-of-the-art performance while enhancing computational efficiency, as demonstrated through extensive zero-shot and fine-tuning evaluations.
Additionally, our training strategy enables a flexible trade-off between inference speed and detail preservation by allowing variable input resolutions at test time while maintaining global scale consistency.}

\section*{Acknowledgments}
This work is funded by Toyota Motor Europe via the research project TRACE-Z\"urich. Additional thanks to Lavinia Recchioni for her editing and unwavering support.




\bibliographystyle{IEEEtran}
\bibliography{IEEEabrv,refs}


\vspace{-2em}

\begin{IEEEbiography}[{\includegraphics[width=1in,clip,keepaspectratio]{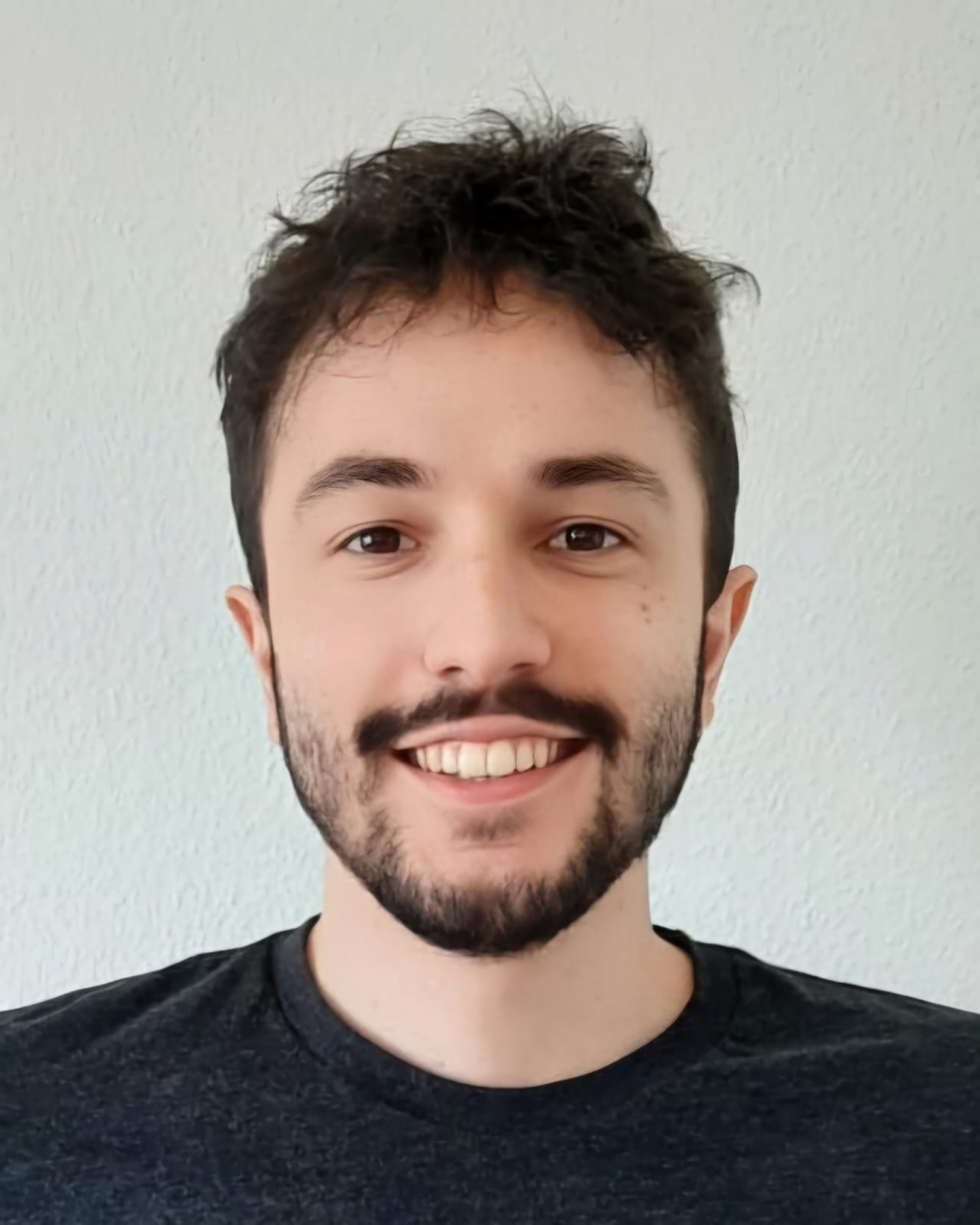}}]{Luigi Piccinelli} is a Ph.D. candidate at ETH Z\"urich, Computer Vision Lab, supervised by Prof. Luc Van Gool and Dr. Christos Sakaridis.
His research focuses on 3D perception, particularly advancing generalization for ill-posed problems such as monocular depth estimation, both from single images and videos.
He has also explored tracking and domain adaptation.
He obtained his B.Sc. and M.Sc. degrees in Electrical Engineering from University of Bologna and ETH Zurich, respectively.
\end{IEEEbiography}

\vspace{-2em}

\begin{IEEEbiography}[{\includegraphics[width=1in,clip,keepaspectratio]{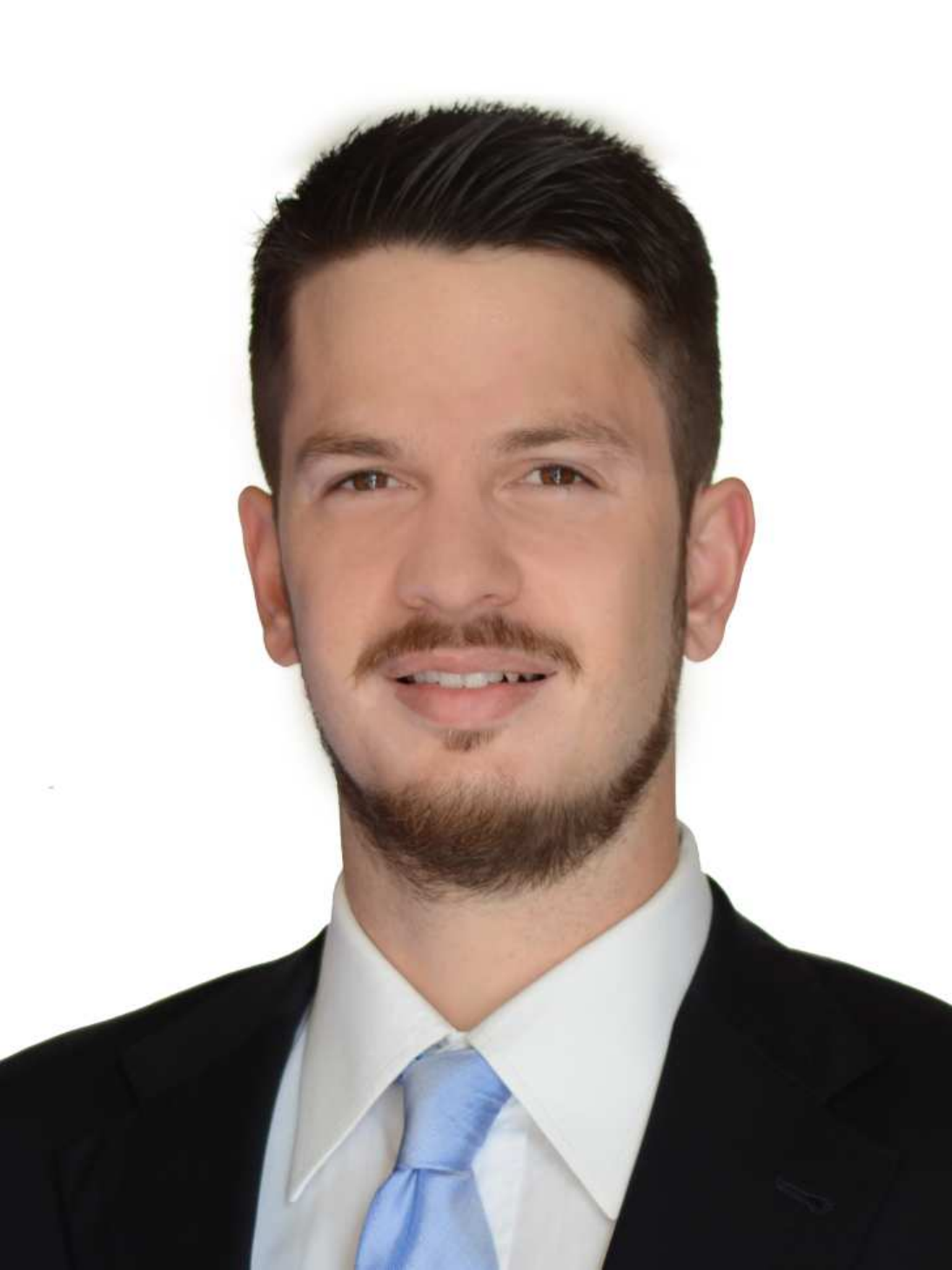}}]{Christos Sakaridis} is a lecturer and Principal Investigator leading the Artificial Visual Intelligence group in the Photogrammetry and Remote Sensing lab of ETH Z\"urich. His research fields are computer vision, artificial intelligence, and machine learning. The focus of his research is on 3D and semantic visual perception, where he develops hybrid, data-driven yet informed, vision models and representations. Since 2021, he leads Toyota TRACE-Z\"urich, a large-scale project on computer vision for autonomous cars and robots. He received the ETH Zurich Career Seed Award in 2022. He obtained his PhD from ETH Z\"urich in 2021, having worked in Computer Vision Lab. Prior to that, he received his MSc in Computer Science from ETH Z\"urich in 2016 and his Diploma in Electrical and Computer Engineering from National Technical University of Athens in 2014.
\end{IEEEbiography}


\begin{IEEEbiography}[{\includegraphics[width=1in,clip,keepaspectratio]{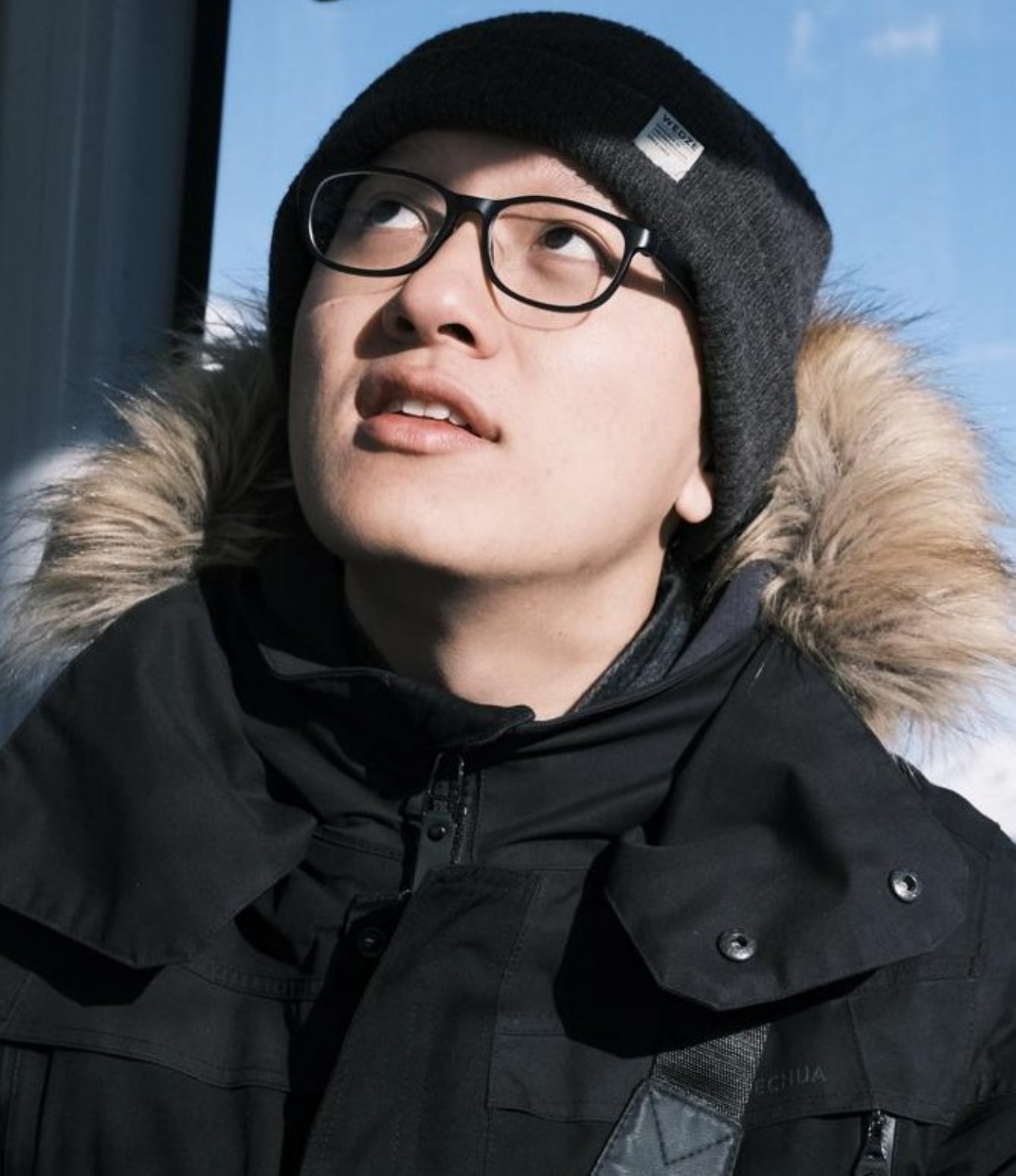}}]{Yung-Hsu Yang} is a Ph.D. student at ETH Z\"urich supervised by Prof. Marc Pollefeys. My research interests include scene understanding and 3D Object Detection and Tracking. He obtained my M.Sc. and B.Sc. degrees in Electrical Engineering at National Tsing Hua University supervised by Prof. Min Sun.
Previously, he worked with Dr. Samuel Rota Bulo and Dr. Peter Kontschieder in dense prediction tasks.
\end{IEEEbiography}

\vspace{-2em}

\begin{IEEEbiography}[{\includegraphics[width=1in,clip,keepaspectratio]{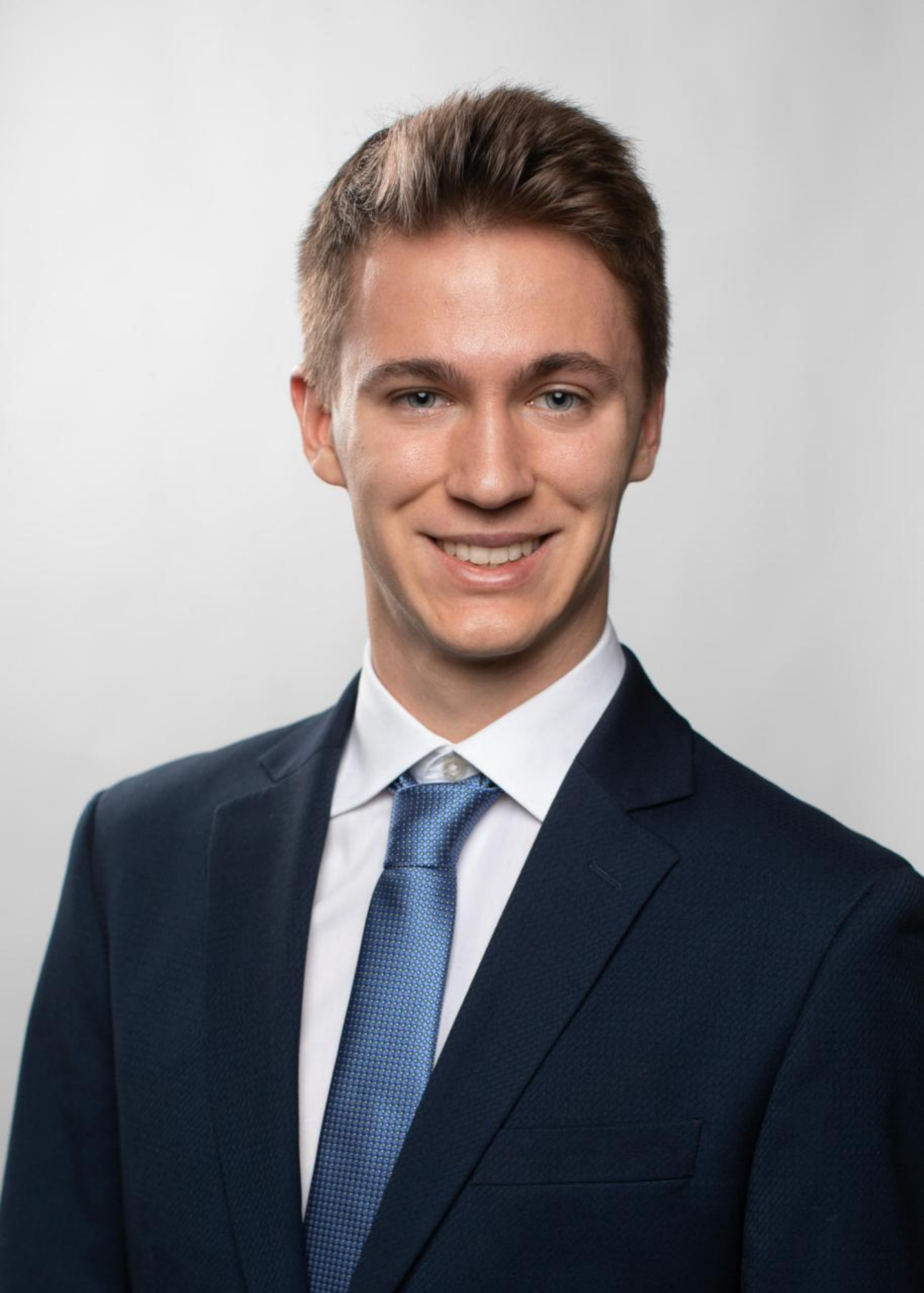}}]{Mattia Segu} is a Ph.D. candidate at the Computer Vision Lab at ETH Z\"urich, co-supervised by Prof. Luc Van Gool and Prof. Bernt Schiele as a member of the Max Planck ETH Center for Learning Systems.
His research focuses on advancing multiple object tracking methods that can learn end-to-end from long video sequences, adapt dynamically, and leverage limited annotations in a self-supervised fashion.
Currently, he is a student researcher at Google, contributing to Federico Tombari's team.
Additionally, he has worked on domain generalization, uncertainty estimation, and foundation models for object tracking and depth estimation.
\end{IEEEbiography}

\vspace{-2em}

\begin{IEEEbiography}[{\includegraphics[width=1in,clip,keepaspectratio]{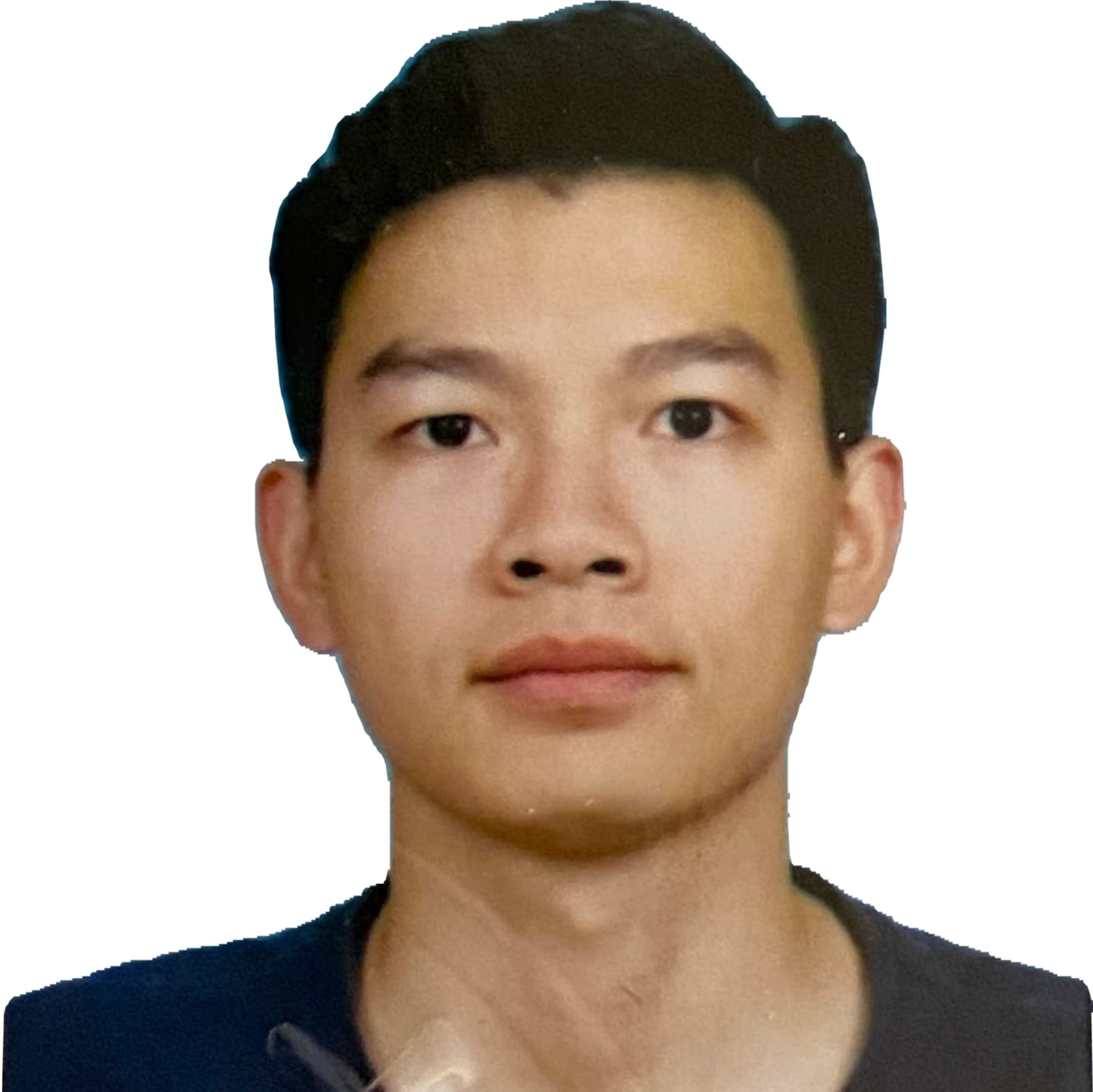}}]{Siyuan Li} is a Ph.D. student at the Computer Vision Laboratory, ETH Z\"urich, Switzerland, supervised by Dr. Martin Danelljan and Prof. Luc Van Gool.
His research focuses on computer vision and machine learning, with an emphasis on visual perception, open-world understanding, and multi-object tracking. He is particularly interested in developing scalable and generalizable models for autonomous driving and robotics.
His work has been published in top-tier computer vision conferences, including CVPR, ECCV, and ICCV.
\end{IEEEbiography}

\vspace{-2em}

\begin{IEEEbiography}[{\includegraphics[width=1in,clip,keepaspectratio]{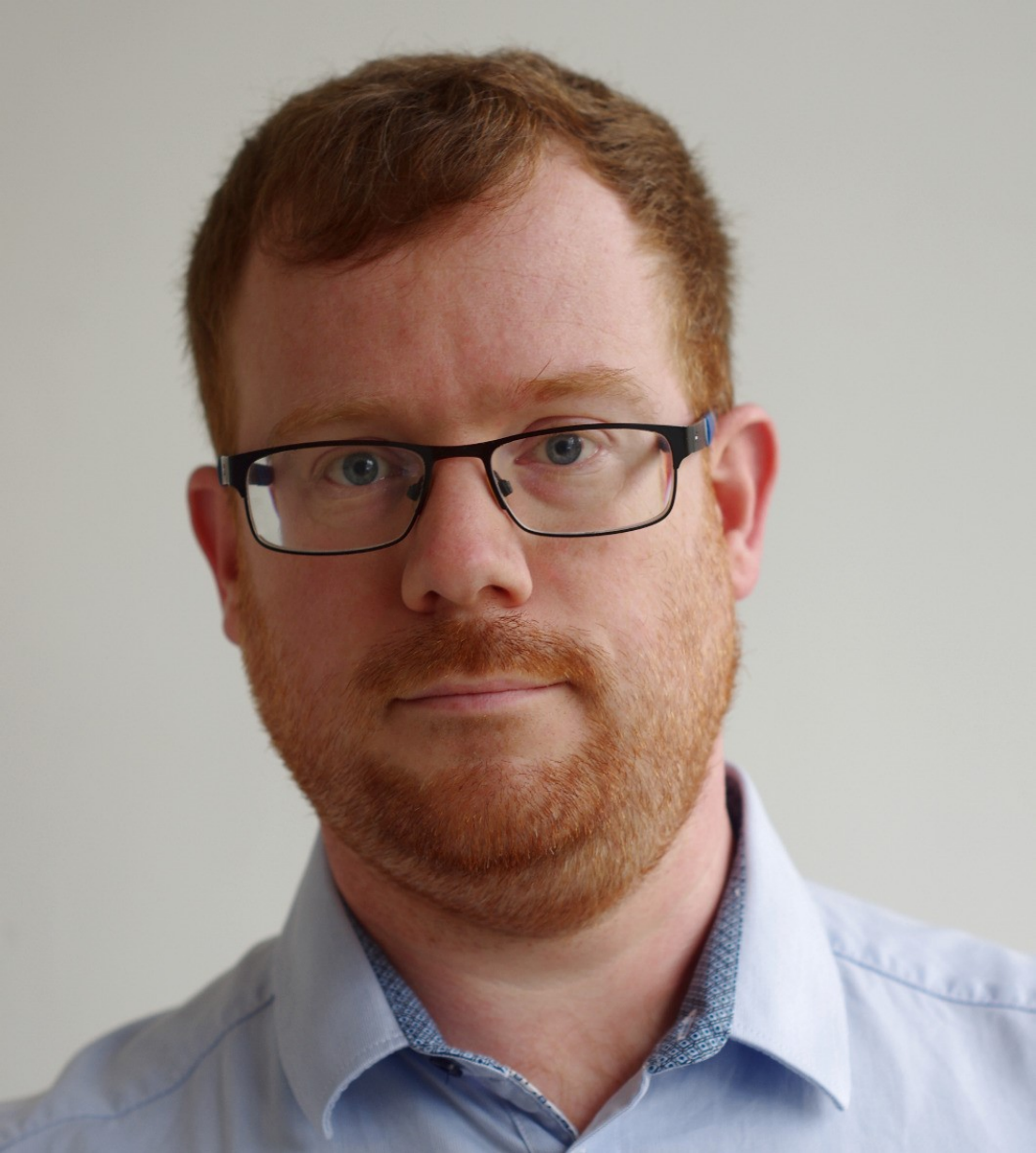}}]{Wim Abbeloos}
He earned an MSc in Applied Engineering from the University of Antwerp (2011) and then worked as a researcher and PhD student at both InViLab (University of Antwerp) and EAVISE (KU Leuven), focusing on 3D object detection, unsupervised 3D object discovery, and pose estimation for robotics. Subsequently, he joined Toyota Motor Europe (Belgium) in 2018, where he currently coordinates and manages research collaborations with top research institutes in Europe in the fields of computer vision and artificial intelligence, including automated driving and other application areas. Additionally, he supports the transfer and integration of the developed knowledge into future applications and products.
\end{IEEEbiography}

\vspace{-2em}

\begin{IEEEbiography}[{\includegraphics[width=1in,clip,keepaspectratio]{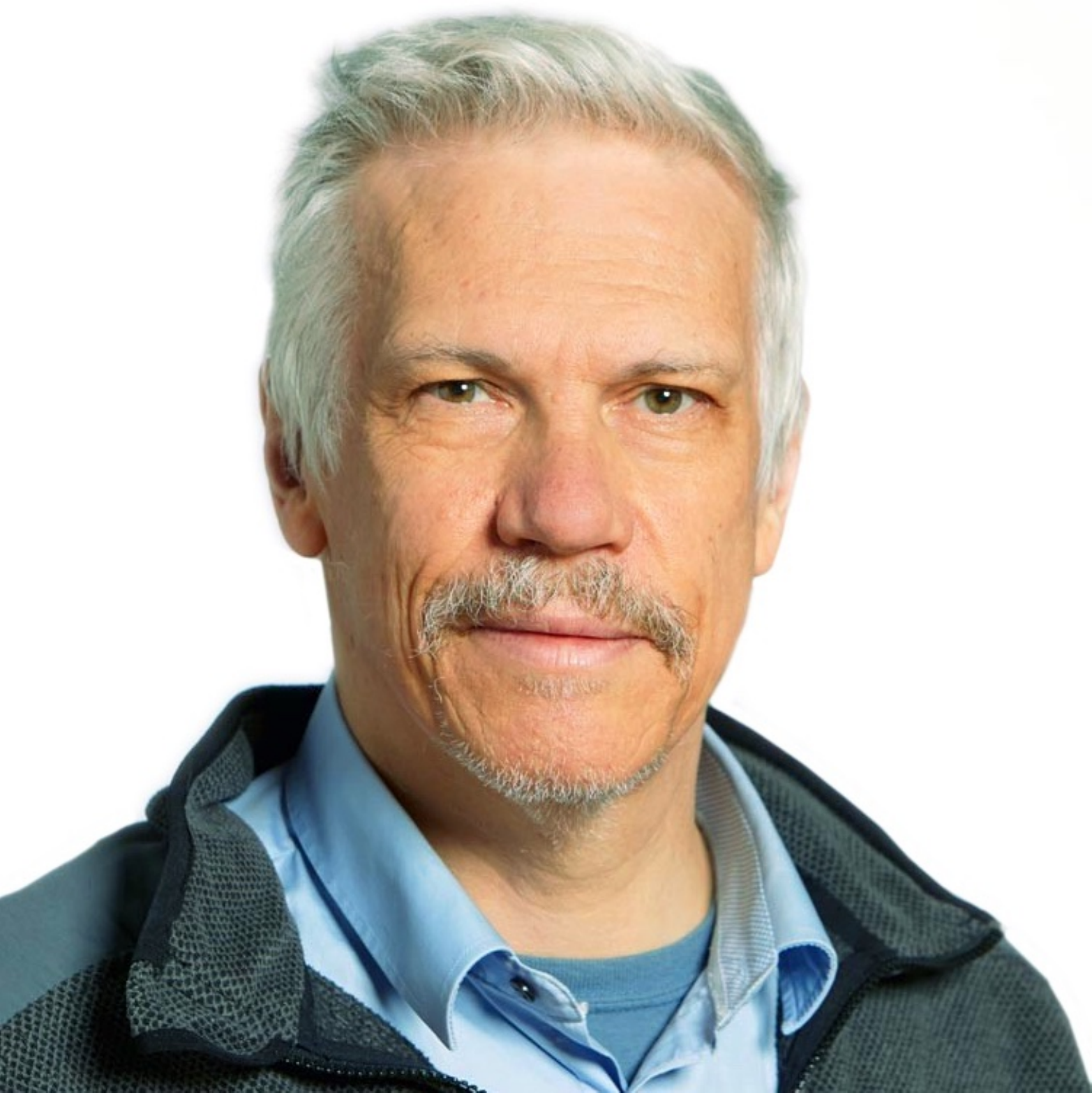}}]{Luc Van Gool} is a full professor for Computer Vision at INSAIT and professor emeritus at ETH Z\"urich and the KU Leuven. He has authored over 900 papers.
He has been a program committee member of several major computer vision conferences (\eg ICCV’05, ICCV’11, and ECCV’14).
His main interests include 3D reconstruction and modeling, object recognition, and autonomous driving.
He received several best paper awards (\eg David Marr Prize ’98, Best Paper CVPR’07). He received the Koenderink Award in 2016 and the ``Distinguished Researcher'' nomination by the IEEE Computer Society in 2017.
In 2015 he also received the 5-yearly Excellence Prize by the Flemish Fund for Scientific Research. He was the holder of an ERC Advanced Grant (VarCity).
Currently, he leads computer vision research for autonomous driving in the context of the Toyota TRACE labs and has an extensive collaboration with Huawei on image and video enhancement.
\end{IEEEbiography}


\end{document}